\documentclass[journal]{IEEEtran}
\usepackage[colorlinks,
            linkcolor=black,       
            anchorcolor=black,  
            citecolor=black,       
            ]{hyperref}

\usepackage{cite}

\usepackage[pdftex]{graphicx}
\usepackage{amsmath}
\usepackage{amssymb}
\usepackage[caption=false,font=normalsize,labelfont=sf,textfont=sf]{subfig}

\usepackage{multirow}

\begin{document}
%
\title{IA-LSTM: Interaction-Aware LSTM for Pedestrian Trajectory Prediction}
%
%
%

\author{Jing~Yang,~\IEEEmembership{Member,~IEEE,}
	    Yuehai~Chen,~\IEEEmembership{Member,~IEEE,}
		Shaoyi~Du,~\IEEEmembership{Member,~IEEE,}\\
		Badong~Chen,~\IEEEmembership{Senior Member,~IEEE,}
		and~Jose C.~Principe,~\IEEEmembership{Fellow,~IEEE,}
\thanks{This work was supported by the National Key Research and Development Program of China under Grant No. 2020AAA0108100, the National Natural Science Foundation of China under Grant No. 62073257, 62141223, and the Key Research and Development Program of Shaanxi Province of China under Grant No. 2022GY-076. (\textit{Corresponding author: Shaoyi Du.})}%
\thanks{Jing Yang and Yuehai~Chen are co-first authors.}%
\thanks{Jing Yang and Yuehai~Chen are with School of Automation Science and Engineering, Faculty of Electronic and Information Engineering, Xi’an Jiaotong, Xi’an 710049, China (e-mail: jasmine1976@xjtu.edu.cn; cyh0518@stu.xjtu.edu.cn.)}%
\thanks{Shaoyi Du and Badong Chen are with Institute of Articial Intelligence and Robotics, College of Articial Intelligence, Xi’an Jiaotong University, Xi’an, Shanxi 710049, China (e-mail: dushaoyi@gmail.com; chenbd@xjtu.edu.cn ).}%
\thanks{Jose C.~Principe is with the Computational NeuroEngineering Laboratory, Department of Electrical and Computer Engineering, University of Florida, Gainesville, FL 32611, USA (e-mail: principe@cnel.ufl.edu).}%
}

%
%

\markboth{Journal of \LaTeX\ Class Files,~Vol.~14, No.~8, August~2015}%
{Shell \MakeLowercase{\textit{et al.}}: Bare Demo of IEEEtran.cls for IEEE Journals}
%



\maketitle

\begin{abstract}
Predicting the trajectory of pedestrians in crowd scenarios is indispensable in self-driving or autonomous mobile robot field because estimating the future locations of pedestrians around is beneficial for policy decision to avoid collision. It is a challenging issue because humans have different walking motions, and the interactions between humans and objects in the current environment, especially between humans themselves, are complex. Previous researchers focused on how to model human–human interactions but neglected the relative importance of interactions. To address this issue, a novel mechanism based on correntropy is introduced. The proposed mechanism not only can measure the relative importance of human–human interactions but also can build personal space for each pedestrian. An interaction module including this data-driven mechanism is further proposed. In the proposed module, the data-driven mechanism can effectively extract the feature representations of dynamic human–human interactions in the scene and calculate the corresponding weights to represent the importance of different interactions. To share such social messages among pedestrians, an interaction-aware architecture based on long short-term memory  network for trajectory prediction is designed. Experiments are conducted on two public datasets. Experimental results demonstrate that our model can achieve better performance than several latest methods with good performance.
\end{abstract}


%
\IEEEpeerreviewmaketitle

\section{Introduction}
%
%
%
%
\begin{figure}
	\centering
	\includegraphics[width=9cm]{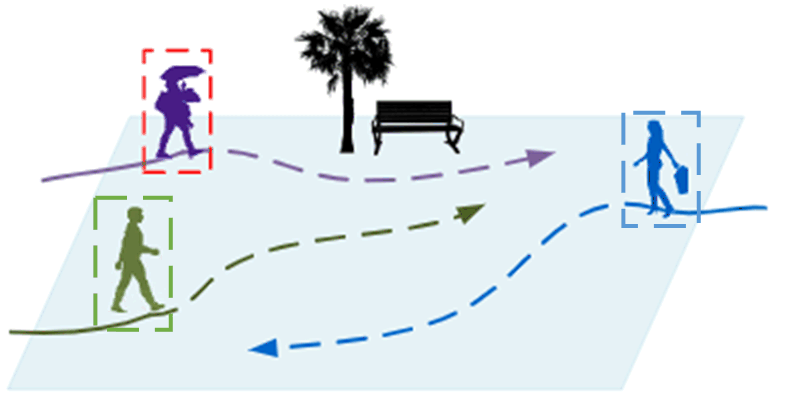}
	\caption{Illustration of a common scene where static interactions and dynamic interactions both occur. The pedestrian who is under an umbrella (framed by dashed red line) has static interactions with stationary obstacles (tree and bench) and dynamic interactions with two other pedestrians, which influence his/her future trajectory.}
	\label{fig1}
\end{figure}

\IEEEPARstart{T}{rajectory} prediction for pedestrians holds significant importance across various applications, including autonomous driving \cite{Lerner2010Crowds}, robot navigation \cite{Luo2018PORCA, Pellegrini2009You, Vivacqua2017Self, Thrun2002MINERVA}, and surveillance camera analysis \cite{Leonard1990Application, Trautman2010Unfreezing}. Accurately anticipating the future positions of pedestrians in the real world is crucial for tasks such as robot-assisted pedestrian regulation \cite{8540942}, person reidentification \cite{8694838}, and target tracking \cite{target_tracking_CYB1, target_tracking_CYB2}. However, predicting pedestrian motions is a highly challenging problem due to the intricate interactions between pedestrians. In real-world scenarios, individuals exhibit behaviors like yielding right-of-way to nearby people \cite{shafiee2021introvert}, walking in groups \cite{moussaid2010walking}, and adhering to social norms, which influence their trajectories based on the movements of other pedestrians \cite{gupta2018social}. One of the primary challenges in trajectory forecasting is effectively integrating human-human interactions into the prediction models for pedestrians.

One of the most classical approaches that considers both static and dynamic interactions is the "Social Force" model \cite{helbing1995social}. This model assumes that objects in the scene, including humans, exert forces on the target pedestrian to influence their movements. These interactions are intelligently modeled by defining various repulsive forces. Similarly, numerous methods have been proposed to consider interaction forces among pedestrians \cite{Lerner2010Crowds, Pellegrini2009You}, vehicles \cite{yang2019top}, and the environment \cite{yamaguchi2011you}. While these methods have achieved competitive results in some datasets, they calculate forces based on handcrafted functions—a fixed form of a physical model simulating changes in behavior. Consequently, these models can only capture simple dynamic interactions and often overlook complex crowd social behaviors \cite{Alahi2016Social}.

Recently, with substantial advancements in deep learning, recurrent neural networks (RNN) and its variants, such as long short-term memory (LSTM) \cite{hochreiter1997long} and gated recurrent unit (GRU) \cite{chung2014empirical}, have become widely utilized. Particularly, they find applications in time series problems, including pedestrian trajectory prediction \cite{Alahi2016Social, Fernando2018Tree, gupta2018social, Kitani2011Fast, Ryoo2015Early, Srivastava2015Unsupervised, vondrick2015anticipating, vemula2018social}.
To account for interactions, social LSTM \cite{Alahi2016Social} introduces a pooling layer to convey interaction information among pedestrians, then applies LSTM to predict human trajectories. It assumes that human-human interactions caused by different pedestrians in the scene are equally important, which might not reflect the real situation comprehensively. In pursuit of enhancing the quality of extracted interaction features, several methods following the social LSTM pattern have been proposed to share information through different mechanisms, such as attention mechanisms \cite{vaswani2017attention} or similarity measures \cite{vemula2018social}. However, these mechanisms introduce additional parameters into the original model. The more pedestrians in the scene, the greater the computational load, significantly escalating training costs.

The pedestrian trajectory prediction task needs to consider human-human interactions to avoid collisions with other pedestrians. It is different from the behavior of other sequence tasks \cite{graves2014towards, graves2013generating}. In the real world, dissimilar interactions occur among different pedestrians. Figure~\ref{fig1} shows a pedestrian (framed by the dashed red line) pays more attention to the interaction caused by a nearby person (framed by the dashed green line), and the interaction caused by the person (framed by the dashed blue line) walking on the distant street is definitely less concerning. However, studying only neighboring pedestrians \cite{Alahi2016Social} is also insufficient to capture dynamic human-human interactions. Inspired by the procedure that the hippocampus processes and integrates spatio-temporal information to form memories, \cite{wu2023multi} propose a novel multi-stream representation learning module to learn complex spatio-temporal features of pedestrian trajectory. \cite{shi2023trajectory} unifies the trajectory prediction components, social interaction, and multimodal trajectory prediction, into a transformer encoder-decoder architecture to effectively remove the need for post-processing. \cite{zhou2023static} proposes a novel global graph representation, which considers the spatial distance (from near to far) and the motion state (from static to dynamic), to explicitly model the social interactions among pedestrians. These methods model the social interactions with neighbors in both temporal and spatial domains. However, these methods ignore that a human has his/her own personal space \cite{Alahi2016Social,personal_space1,personal_space2,personal_space3,personal_space4}. Humans respect personal space: Once a pedestrian enters the personal space of other people, the generated interaction is large, but it drops sharply outside this personal space. This type of human-human interactions cannot be measured just by the relative Euclidean distance between pedestrians, which is the most common measurement method in recent research \cite{yamaguchi2011you, golchoubian2023pedestrian}. The accurate measurement of human-human interactions is the key issue of pedestrian trajectory prediction \cite{zhang2023dual, 10103218}.

As discussed above, diverse interactions among pedestrians and dynamic human-human interactions exhibit a unique characteristic related to everyone's personal space. To address the aforementioned issues, this paper proposes a novel mechanism based on correntropy. The suggested mechanism aims to measure the relative importance of human-human interactions and represent each individual's personal space. In this mechanism, the correntropy value between two pedestrians is calculated and utilized as the weight of their interactions. Personal space is defined by selecting a suitable correntropy kernel.
Subsequently, we introduce an "Interaction Module" designed to extract feature representations of human-human interactions and assign weights to them. Finally, the social information arising from these human-human interactions is shared through a variant structure of LSTM that we have devised.

Correntropy is computed directly from original data, eliminating the need for hand-crafted functions with specific settings. Its capability to generalize high-dimensional features, incorporating both temporal structure and statistical distribution, makes correntropy well-suited for measuring human-human interactions in pedestrian trajectory prediction tasks. Our designed Interaction-Aware LSTM (IA-LSTM) architecture autonomously captures various complex human-human interactions among crowds through the proposed "Interaction Module," without relying on additional notations or predefined social rules.
This Interaction-Aware LSTM is adept at discerning intricate human-human interactions, and the acquired information is shared among pedestrians in the scene for subsequent trajectory prediction. The efficacy of our approach is validated through experiments on two public datasets, ETH \cite{pellegrini2010improving} and UCY \cite{Lerner2010Crowds}. The experimental results demonstrate that our model outperforms several state-of-the-art methods, showcasing its effectiveness. Additionally, an analysis of the prediction results provides insights into human-human interactions in a social context.

The main contribution of this article is three-fold.

\begin{enumerate}
    \item To quantify the relative importance of human–human interactions and establish personal space for each pedestrian, we introduce a novel mechanism based on correntropy.

    \item To adeptly extract feature representations of dynamic human–human interactions in the scene and compute corresponding weights that signify the importance of different interactions, we propose an interaction module that incorporates a data-driven mechanism.

    \item Experimental results demonstrate that our model outperforms several state-of-the-art methods, showcasing its strong performance.
\end{enumerate}

The reminder of this paper is as follows. In Section 2, a brief introduction of pedestrian trajectory prediction methods, correntropy, and LSTM network methods are provided. In Section 3, the proposed model is detailed. In Section 4, the performance of our model approach is verified, and the predicted trajectories are analyzed to demonstrate the behavior of pedestrians learned by our model. In Section 5, a conclusion is made on our proposed methods.

\section{Related Works and Preliminary}
\subsection{Methods for trajectory prediction of pedestrians}

\noindent \textbf{Traditional Methods.}
Some previous traditional researchers mainly focus on how to model the real human among the crowds and use this descriptive model to predict future trajectories. For instance, some early works were based on the Bayesian model, using online Bayes filters (Kalman filters and particle filters) \cite{Schneider2013Pedestrian} or a dynamic Bayesian network \cite{Ballan2016Knowledge, kooij2014context-based}. These models make predictions without considering whether static or dynamic interactions are in the scene, and the predicted results highly deviate from the ground truth.

Another kind of traditional approaches consider static interactions (static texture) in the scene using a grid graph to predict entire future positions at the same time. These approaches design a grid graph to represent all possible paths of one pedestrian and assign different weights to edges; therefore, the prediction problem becomes the shortest path problem. In particular, Huang \textit{et al.} \cite{huang2016deep} and Walker \textit{et al.} \cite{walker2014patch} used the appearance texture of the scene to build the cost map. Xie \textit{et al.} \cite{Dan2013Inferring} used objects in the scene to define the cost map. 
There are other traditional methods for trajectory prediction task. Yamaguc \textit{et al.} \cite{yamaguchi2011you} viewed pedestrians as decision-making agents that also consider a plethora of personal, social, and environmental factors to decide where to go next. Yang \textit{et al.} \cite{yang2019top} built two datasets for pedestrian motion models that consider both interpersonal and vehicle crowd interaction. With tracking multiple people in complex scenarios, a powerful dynamic model \cite{Pellegrini2009You} is still competitive in some datasets nowadays \cite{Lerner2010Crowds}. However, these works mostly use hand-crafted functions, a fixed form of physical model, to calculate changeable interaction forces. The models could only capture simple dynamic interactions and might fail to generalize for more complex scenes \cite{Alahi2016Social}. The inadequacy of existing approaches to account for dynamic human-human interactions renders them unsuitable for crowd scenarios. In contrast, our proposed 'Interaction Module' is designed to extract real-time feature representations of dynamic human-human interactions, thereby enhancing the accuracy of pedestrian trajectory prediction.

\noindent \textbf{Deep Learning Methods.}
Recently, with enormous advancements of deep learning, numerous methods are applied to address this task and achieve better results. For instance, convolutional neural networks (CNNs) were applied by Yi \textit{et al.} \cite{Shuai2016Pedestrian} to make predictions. They proposed a named behavior-CNN that has three dimensional data channels with bias maps to consider different behaviors at specific locations such as entrances and obstacles in the scene. With the limitation of input form, this method only considers static interactions in the scene while neglecting the influence caused by other pedestrians.

In addition, Inverse Reinforcement Learning (IRL) \cite{Abbeel2016Inverse, Ng2000Algorithms} was first presented to solve the optimal motion planning of robots \cite{Ziebart2009Planning}, and Kitani \textit{et al.} \cite{Kitani2012Activity} cleverly introduced IRL to trajectory prediction. Instead of directly predicting the future path, they taught the robot to take actions sequentially and obtain its entire future positions as the predicted path. Lee \textit{et al.} \cite{Lee2016Predicting} also applied this method to predict the moving trajectory of the football player and Wei \textit{et al.} \cite{ma2017forecasting} proposed an approach based on fictitious play theory to predict trajectories of multiple pedestrians with several predefined goals. In general, these works mentioned above are named as activity forecasting. This kind of methods converts static and dynamic interactions to what the target robot observes, but the most difficult part is how to define the reward, namely, the effect of these different interactions. In their work, \cite{bae2023set} introduces a trajectory prediction model based on graph convolutional networks. This model incorporates a control point prediction strategy, dividing future paths into three sections and inferring pedestrians' intermediate destinations to mitigate cumulative errors. Similarly, \cite{liang2023stglow} puts forth a pioneering generative flow-based framework featuring a dual-graphormer. This framework enhances the precision of modeling underlying data distribution by optimizing the exact log-likelihood of motion behaviors.
While these approaches demonstrate impressive performance in trajectory prediction, they overlook a crucial aspect—the influence of human personal space. Human-human interaction dynamics undergo significant changes based on proximity, a factor not considered in their methodologies. In contrast, our proposed mechanism, rooted in correntropy, uniquely captures everyone's personal space and evaluates the relative importance of human-human interactions. This novel approach proves effective in enhancing trajectory prediction tasks.

Numerous commendable studies have delved into trajectory prediction, each offering unique perspectives. Following object detection for location determination, \cite{CYB_2023_1} introduces a Time Profit Elman Neural Network (TPENN) dedicated to forecasting the trajectory of dynamic targets. In a different vein, \cite{CYB_2023_2} employs a game-theoretic driver steering control model to characterize human drivers' steering responses to automated interventions. Meanwhile, \cite{CYB_2023_3} devises an Event-Triggered Model Predictive Control (EMPC) strategy, ensuring trajectory tracking and obstacle avoidance. The work of \cite{CYB_2023_4} puts forth a Variable Centroid Trajectory Tracking Network, fusing vision and force information to achieve highly precise prediction and tracking. Additionally, \cite{CYB_2023_5} introduces a deep learning framework featuring a Spatiotemporal Attention Network with a Neural Algorithm Logic Unit, specifically designed to comprehend the dynamic aggregation effects of private cars on weekends. These methodologies predominantly focus on the object trajectory prediction task, where measuring interactions proves to be simpler compared to the complexities involved in predicting pedestrian trajectories.

\subsection{Correntropy}
Correntropy, the abbreviation for correlation entropy, proposed by Principe \textit{et al.} \cite{Liu2007Correntropy, Principe:2010:ITL:1855180}, is an effective kernel-based similarity measure in feature space \cite{Santamar2006Generalized} based on information theoretic learning \cite{Principe2010Information} with various applications such as classification \cite{ShiTraining, Singh2014The, Syed2012Correntropy}, regression \cite{Chen2012Recursive, Feng2015Learning}, deep learning \cite{ShiTraining, Chen2016Efficient}, pitch detection in speech \cite{Xu2008A}, adaptive filtering \cite{Chen2014Steady, Zhao2011Kernel}, and principal component analysis \cite{Chen2018Robust, Ran2011Robust}. 

Humans inherently respect personal space and adhere to the practice of yielding right-of-way. Each pedestrian maintains a comfortable personal space, and discomfort arises when others encroach upon this designated area. The influence of generated interactions is significant within the confines of personal space, sharply diminishing outside its boundaries. Furthermore, a pedestrian is more susceptible to the influence of nearby individuals compared to those at a distance. In this context, distant pedestrians can be regarded as outliers. 

We extend correntropy to trajectory prediction to establish personal space for each pedestrian and gauge the relative importance of human-human interactions in the scene. Specifically, we initially construct the personal space of pedestrians by designing a suitable Gaussian kernel. The constructed personal space contains the robustness to outliers, which is leveraged to model human-human interactions.  In comparison to other similarity measures, such as mean-square error, our method based on correntropy demonstrates the capability to generalize the conventional correlation function to high-dimensional feature spaces and exhibits enhanced resilience to impulsive noises or outliers \cite{Chen2018Robust, chen2021multikernel, 8303753, 7447745, 8959408}. Our method based on correntropy defines a non-homogeneous metric, making it an effective outlier-robust error measure in robust signal processing \cite{chen2021multikernel}. Similarly, Du \textit{et al.} \cite{8303753} proposed a robust graph-based method based on correntropy to handle labeling noisy data. The constructed personal space based on correntropy, calculated directly from original data without additional settings, is inherently suitable for such pedestrian trajectory prediction scenarios.


\subsection{Vanilla Long Short-Term Memory Networks}

LSTM is a recurrent neural network variant proposed by Sepp Hochreiter and Jürgen Schmidhuber \cite{hochreiter1997long} to solve the gradient vanishment or gradient explosion issue in conventional back-propagation through time \cite{M1995BTT} and real-time recurrent learning \cite{Williams1989Experimental}. 
Unlike traditional RNNs \cite{mikolov2010recurrent}, an LSTM network is well-suited to learn from experience to classify, process and predict time series given very long time lags. LSTM has various applications in dealing with the time series problem such as natural language process \cite{Chorowski2014End, chung2015a, Graves2013Speech, young2017recent,LSTM1, LSTM2,LSTM3} including sentiment analysis \cite{LSTM1}, air-quality prediction \cite{LSTM3}, image generation \cite{denoord2016pixel, Karpathy2015Deep, LSTM_CV}, fault diagnosis \cite{LSTM4}, and machine translation \cite{sutskever2014sequence}.

Inspired by the recent success of LSTMs in different sequence prediction tasks, most works extend them for pedestrian trajectory prediction. Social LSTM \cite{Alahi2016Social} applies LSTM in human trajectory prediction task with a designed “social” pooling layer for learning typical human-human interactions. Group-LSTM \cite{Niccol2018Group} exploits motion coherency to cluster trajectories having similar motion trends, and then applies an improved LSTM to estimate the future path prediction. SR-LSTM \cite{zhang2019sr} proposes a data-driven state refinement module for LSTM network, which could activate the utilization of the current intention of neighbors. To introduce vital velocity features, CF-LSTM \cite{2020CF} designs a novel feature-cascaded framework for LSTM. These methods improve the quality of extracted interaction features, but they neglect the fact that humans would respect personal space and yield right-of-way. In comparison, correntropy mechanism is adopted to represent everyone’s personal space and measure the relative importance of human-human interactions.

Given the robustness of correntropy to impulsive outliers, we have devised a novel mechanism utilizing correntropy to precisely model interactions among individuals in crowded scenarios. This correntropy-based mechanism is adept at representing the personal space of each individual and quantifying the relative importance of human-human interactions. Furthermore, to seamlessly integrate accurate human-human interactions into the trajectory prediction model, we introduce an "Interaction Module" designed to autonomously share these interactions among all pedestrians in the scene.

\begin{figure*}[htbp]
	\centering
	\includegraphics[width=17cm]{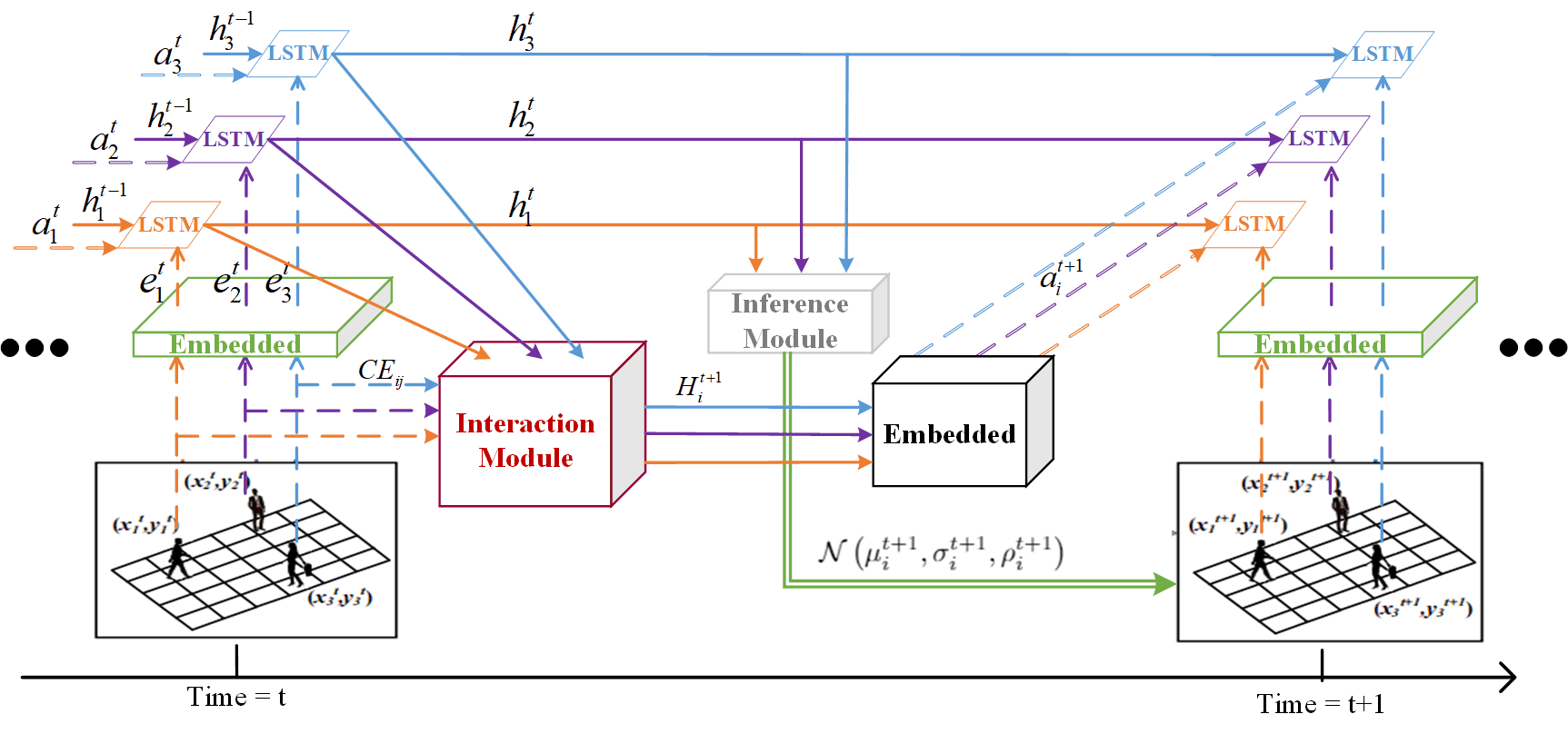}
	\caption{Overview of our model.}
	\label{fig3}
\end{figure*}

\section{Methodology}

Pedestrians within a scene dynamically adjust their behaviors in response to others, with various interactions eliciting distinct reactions. Inspired by this, we introduce a model capable of comprehending human-human interactions while simultaneously predicting trajectories.

This section offers a comprehensive overview of our innovative LSTM structure, complemented by an "Interaction Module," designed to predict trajectories across the entire scene. Our complete proposed model is denoted as the IA-LSTM model.

\subsection{Problem Definition}
First, the frames of a video with a fixed interval are preprocessed to obtain the spatial coordinates of each pedestrian. The coordinates $(x_{i}^{t},y_{i}^{t})\in \mathbb{R}^2$ of pedestrian $i$ at time $t$ are denoted as $\vec{p}_{i}^{t}$. Then, the trajectory prediction problem is formally described as follows. The future trajectory $\Gamma_i=(\vec{p}_{i}^{obs+1},...,\vec{p}_{i}^{pred})$ of target pedestrian $i$ from time steps $t=T_{obs+1},...,T_{pred}$ is predicted, considering his/her own past trajectory $\mathcal{H}_i=(\vec{p}_{i}^{1},...,\vec{p}_{i}^{obs})$ from time steps $t=1,...,T_{obs}$ and other pedestrians' trajectory information in the scene $\{\mathcal{H}_j:j \in {1,2,...,N}, j \neq i\}$, where $N$ denotes the number of pedestrians in the scene.

Our goal is to learn the parameters $W^{*}$ of a model to predict the future locations of each pedestrian between $t=T_{obs+1}$ and $t=T_{pred}$. Formally,
\begin{equation}
\Gamma_i = f(\mathcal{H}_1,\mathcal{H}_2,...,\mathcal{H}_N;W^{*})
\label{}
\end{equation}
where $W^{*}$ is the collection of all parameters used in our model $f$. The details are elaborated in the following section.

\subsection{Model Description}
As mentioned above, trajectory prediction can be considered a time series prediction problem, and LSTM network is well suited for this task \cite{hochreiter1997long}. In previous work, each vanilla LSTM represents one pedestrian, the spatial coordinates at some point are used as the input, and the hidden states of vanilla LSTM are used to estimate the future locations of pedestrians. In this vanilla LSTM, a fatal weakness is that each pedestrian is modeled independently by a separate LSTM, and dynamic human-human interactions cannot be shared among pedestrians. Pedestrians are isolated in this framework, which makes them unable to interact with people around them, and the prediction error is large \cite{Alahi2016Social}. To address this issue, a variant structure of LSTM is designed, as illustrated in Figure \ref{fig3}. In this structure, each LSTM still represents one pedestrian, but an ``Interaction Module" for passing or sharing dynamic human-human interactions among pedestrians in the scene is introduced. In our framework, each pedestrian can receive interaction information from people around him/her.

First, as shown in the green ``Embedded Module" of Figure \ref{fig3}, at time $t$, the coordinates of the $i^{th}$ pedestrian $(x_{i}^{t}, y_{i}^{t})$ are embedded into a vector $\boldsymbol{e}$ as follows:
\begin{equation}
e_{i}^{t}=\phi\left(x_{i}^{t}, y_{i}^{t} ; W_{e}\right)
\label{eq3}
\end{equation}
where $\phi$ is the embedding function with ReLU nonlinearity, and $W_e$ are the embedding parameters.  

Then, to share the information with one another, an ``Interaction Module" trying to extract different relations among pedestrians is defined. The red ``Interaction Module" shows an ``Interaction Tensor" $H_{i}^{t}$ as the output of the ``Interaction Module" is constructed to represent human-human interactions occurring around target pedestrian $i$ at time $t$. For sharing the relations, similarly, ``Interaction Tensor" $H_{i}^{t}$ is embedded into vector $\boldsymbol{a}$ as follows.
\begin{equation}
a_{i}^{t}=\phi\left(H_{t}^{i} ; W_{a}\right)
\label{eq4}
\end{equation}
where $\phi$ is the embedding function with ReLU nonlinearity, and $W_a$ are the embedding parameters.

Vector $\boldsymbol{e}$ represents the spatial feature information of the target pedestrian, and vector $\boldsymbol{a}$ represents the feature information of human-human interactions acting on the target pedestrian in the scene. Finally, they are concatenated as one input to the LSTM cell, which is formulated as follows.
\begin{equation}
h_{i}^{t}=\operatorname{LSTM}\left(h_{i}^{t-1}, concat(e_{i}^{t}, a_{i}^{t}) ; W_{l}\right) 
\label{eq5}
\end{equation}
where $W_l$ are the LSTM parameters, and all LSTM parameters are shared across pedestrians at the same time step. 

\subsubsection{Inference Module}
In the inference module, a bivariate Gaussian distribution parameterized by mean $\mu_{i}^{t}=\left(\mu_{x}, \mu_{y}\right)_{i}^{t}$, standard deviation $\sigma_{i}^{t}=\left(\sigma_{x}, \sigma_{y}\right)_{i}^{t}$, and correlation coefficient $\rho_{i}^{t}$ is assumed to estimate the predicted coordinates. These parameters at time $t+1$ are determined by hidden state $h_{i}^{t}$ at time $t$ passing through a linear layer $W_o$ as follows:
\begin{equation}
\left(\mu_{i}^{t+1}, \sigma_{i}^{t+1}, \rho_{i}^{t+1}\right)=W_{o} h_{i}^{t}
\label{eq6}
\end{equation}

The predicted coordinates are given by the following:
\begin{equation}
(\hat{x}_{i}^{t+1}, \hat{y}_{i}^{t+1}) \sim \mathcal{N}\left(\mu_{i}^{t+1}, \sigma_{i}^{t+1}, \rho_{i}^{t+1}\right)
\label{eq7}
\end{equation}

Our model is jointly trained by minimizing the negative log-likelihood loss $L_i$ ($L_i$ represents the $i^{th}$ trajectory) as follows:
\begin{equation}
L^{i}\left(W_{e}, W_{a},W_{l}, W_{o}\right)=-\sum_{t=T_{obs}+1}^{T_{pred}} \log \left(\mathbb{P}\left(x_{i}^{t}, y_{i}^{t} | \sigma_{i}^{t}, \mu_{i}^{t}, \rho_{i}^{t}\right)\right)
\label{eq8}
\end{equation}

The loss is calculated over the entire trajectories in the training datasets. Joint back propagation is done through our model at every time step and tuning the parameters to minimize the loss.

\subsubsection{Interaction Module}
Pedestrians exhibit diverse walking motion patterns, each possessing distinct velocities, orientations, and goals. They inherently respect personal space, adapting their actions based on real-time interactions with the surrounding environment. As highlighted earlier, human-human interactions play a pivotal role in predicting pedestrians' next moves. Some existing methods, such as \cite{Luber2010People, gupta2018social, helbing1995social, Alahi2016Social, vemula2018social, pellegrini2010improving, Mehran2009Abnormal}, consider these interactions to be equally important or employ specific settings, using simplistic "repulsion" or "attraction" functions to gauge their significance. However, such approaches may be unsuitable for trajectory prediction, particularly in crowded scenarios. For instance, a person entering the personal space of the target pedestrian undoubtedly induces more interactions than someone who is distant from the target pedestrian.

\begin{figure}
	\centering
	\includegraphics[width=8cm]{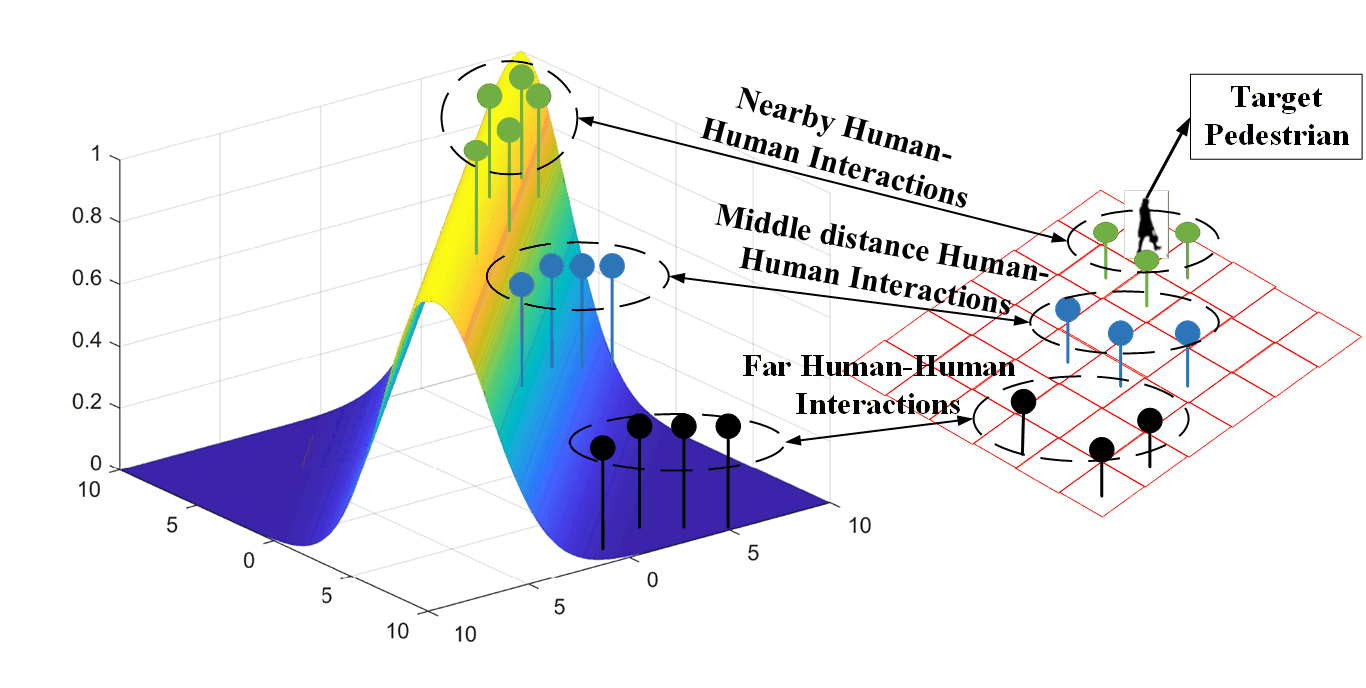}
	\caption{Calculation of correntropy between the target pedestrian and the pedestrians around him/her.}
	\label{figc}
\end{figure}

As a robust nonlinear similarity measure, correntropy is naturally suitable in such pedestrians trajectory prediction case. Correntropy has the ability to build personal space for each pedestrian through setting a suitable Gaussian kernel. In Figure~\ref{figc}, some pedestrians (green representation) enter the personal space of the target pedestrian, which greatly effects the target pedestrian, and the generated interactions are evidently immense.
Moreover, correntropy is very robust to impulsive noises or outlier points owing to the properties of Gaussian kernel \cite{Chen2018Robust, chen2021multikernel, 8303753, 7447745, 8959408,  9044739, 9511644, 9475523, 9115431}. Hence, in this case, human-human interactions caused by pedestrians far from the target pedestrian, the black representations shown in Figure~\ref{figc}, are considered outliers and attenuate the effects. 
Considering the great advantages of correntropy, first, the correntropy between the target pedestrian and the pedestrians around him/her is calculated. 
The correntropy representing human-human interaction between the $i^{th}$ pedestrian and the $j^{th}$ pedestrian is given by the following:
\begin{equation}
CE_{ij}=\exp(-\|\vec{p}_{i}^{t}-\vec{p}_{j}^{t}\|^{2}/ 2\sigma^2)
\label{eq9}
\end{equation}
where $\vec{p}_{i}^{t}$ and $\vec{p}_{j}^{t}$ are the spatial coordinates $(x^t,y^t)$ of the $i^{th}$ and the $j^{th}$ pedestrian at time $t$, respectively; $\|\vec{p}_{i}^{t}-\vec{p}_{j}^{t}\|^{2}$ calculates the Euclidean distance between the $i^{th}$ and the $j^{th}$ pedestrian. The value of correntropy (between $0$ and $1$) represents the relative importance of their interactions ($1$ is the most, and $0$ is the least).

\begin{figure}
	\centering
	\includegraphics[width=8cm]{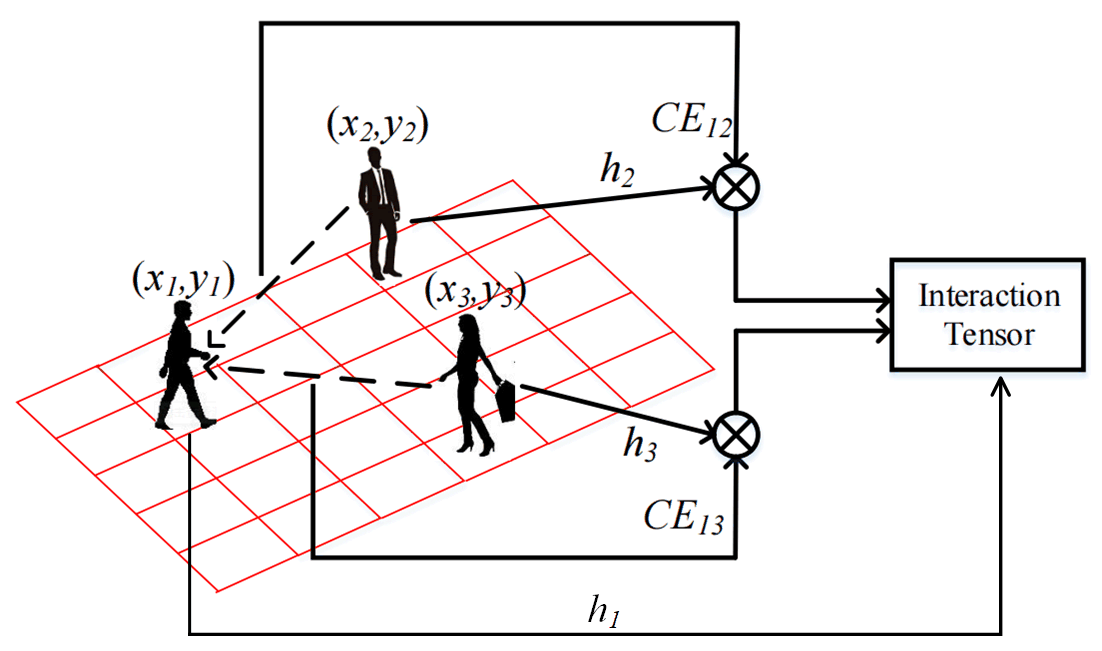}
	\caption{Diagram of the Interaction Module.}
	\label{fig4}
\end{figure}

Then, the obtained relative importance of human-human interactions, is used to construct an ``Interaction Tensor", which represents human-human interactions among crowds in the scene. The target pedestrian ($x_{1}, y_{1}$) shown in Figure \ref{fig4} has his/her own walk state and would be influenced by other pedestrians. The trajectory of the target pedestrian is jointly determined by his/her own walk state and interactions generated by other pedestrians.
First, the feature representations $h_{i}^{t}$ of other pedestrians in the scene are extracted. In LSTM network, hidden state $h_{i}^{t}$ at time $t$ captures the latent representation of the $i^{th}$ pedestrian at that instant. Then, the value that represents the relative importance of the human-human interaction and the hidden state are multiplied to construct an ``Interaction Tensor" for sharing feature information among crowds in the scene. More specifically, the “Interaction Tensor” $H_{i}^{t}$ of the $i^{th}$ pedestrian including itself hidden state $h_{i}^{t-1}$ and interactions from other pedestrians at time $t-1$ are constructed as follows:
\begin{equation}
H_{i}^{t}(x_i,y_i,:)=h_{i}^{t-1} + \sum_{j=1,j\neq i}^{M_i} CE_{ij} \cdot h_{j}^{t-1} =\sum_{j=1}^{M_i} CE_{ij} \cdot h_{j}^{t-1} 
\label{eq10}
\end{equation}
where $h_{j}^{t-1}$ is the hidden state of the LSTM representing the $j^{th}$ pedestrian at time $t-1$, and $M_i$ is the set of other pedestrians who have interactions with target pedestrian $i$ in coordinate $(x_i,y_i)$.

\section{Experiments and Analysis}
In this section, the experimental results of our approach are demonstrated in two public datasets: ETH \cite{pellegrini2010improving} and UCY \cite{lealtaixe2014learning}.
The ETH dataset has 750 pedestrians and two scenes (ETH and Hotel). The UCY dataset has 786 pedestrians and three scenes (ZARA01, ZARA02 and UCY). These two datasets are collected from the real world, containing complex situations such as pedestrians walking in groups, non-linear trajectories with different velocities, intentional avoidance of collisions, and other challenging behaviors, which is suitable for our experiments.

\textbf{Evaluation Metrics:} Similar to \cite{Pellegrini2009You,Alahi2016Social,  vemula2018social}, two following metrics are used. Assuming $N$ is the number of trajectories in testing, $\vec{p}_{i,pred}^{t}$ represents the predicted spatial coordinates $(x^t,y^t)$ of the $i^{th}$ pedestrian at time $t$, and $\vec{p}_{i,obs}^{t}$ represents the observed location.
\begin{itemize}
	\item \textit{Average Displacement Error (ADE):} Introduced in \cite{Pellegrini2009You}, this error calculates the mean distance between all predicted points and actual points in one trajectory. 
	\begin{equation}
	ADE=\frac{\sum_{i=1}^{N} \sum_{t=T_{obs}+1}^{T_{pred}}\left(\vec{p}_{i,pred}^{t}-\vec{p}_{i,obs}^{t}          \right)^{2}}{N\left(T_{pred}-\left(T_{obs}+1\right)\right)}
	\label{eq_ADE}
	\end{equation}
	\item \textit{Final Displacement Error (FDE):} Introduced in \cite{Alahi2016Social}, this error calculates the mean distance between the final predicted point and the final actual point at the end of prediction $T_{pred}$.
	\begin{equation}
	FDE=\frac{\sum_{i=1}^{N} \sqrt{\left( \vec{p}_{i,pred}^{t}-\vec{p}_{i,obs}^{t} \right)^{2}}}{N}
	\end{equation}
\end{itemize}

\textbf{Baselines:} The ``Social LSTM" model and the ``Social Attention" model are selected as our baselines for comparison. 
\begin{itemize}
	\item \textit{Social LSTM Model}: The ``Social LSTM" model outperforms the linear model, the ``Social Force" model \cite{helbing1995social}, and the Interacting Gaussian Processes model \cite{Trautman2010Unfreezing}.
	\item \textit{Social Attention Model}: The ``Social Attention" model outperforms the ``Social LSTM" model on some datasets. 
\end{itemize}

\textbf{Implementation Details:} 
During training, a leave-one-out approach, where our model is trained and validated on four sets and tested on the remaining one, is used. Our baselines, the ``Social LSTM" model and the ``Social Attention" model, are trained in the same manner. During testing, the trajectory for eight frames is observed and the next 12 frames are predicted. The frame rate is $0.4$, which means $T_{obs}=3.2secs$, $T_{pred}-T_{obs}=4.8secs$. It is also the same in our baselines. 

The ``Interaction Tensor" size is $N_{o} \times N_{o} \times D $ , where $N_{o}$ is 128, $D$ is the dimension of the hidden state, and $D$ is set as 128 for all the LSTM models. All the inputs are embedded into 64 dimensional vectors with ReLU nonlinearity. The batch size is 8, and the model is trained for 150 epochs using Adam with an initial learning rate of 0.001. These settings are the same with the baselines. Importantly, different values of $\sigma$ are set to study the performance. 

During testing, our trained model is used to determine the parameters of the bivariate Gaussian distribution and then sample from it to obtain the coordinates $(\hat{x}, \hat{y})_{i}^{t}$ of the $i^{th}$ pedestrians according to Equation \eqref{eq7}. To reduce random errors, each test is run for 50 times, and the averages and variances are calculated as the final results.

From time $T_{obs+1}$ to $T_{pred}$, the actual coordinates $\left(x_{i}^{t}, y_{i}^{t}\right)$ in Equations \eqref{eq3}, \eqref{eq4}, and \eqref{eq5} are replaced with the predicted coordinates $(\hat{x}_{i}^{t}, \hat{y}_{i}^{t})$ to make predictions. Moreover, the predicted coordinates are used to calculate the “Interaction Tensor” in Equations \eqref{eq9} and \eqref{eq10}.

The whole model is trained on a single NVIDIA 3090 GPU with a PyTorch implementation, and all the results as well as variances are shown in Table \ref{tab1}.
\subsection{Performance Study of $\sigma$}
\begin{figure*}	
	\centering	
	\subfloat[ADE]{\includegraphics[width=8cm]{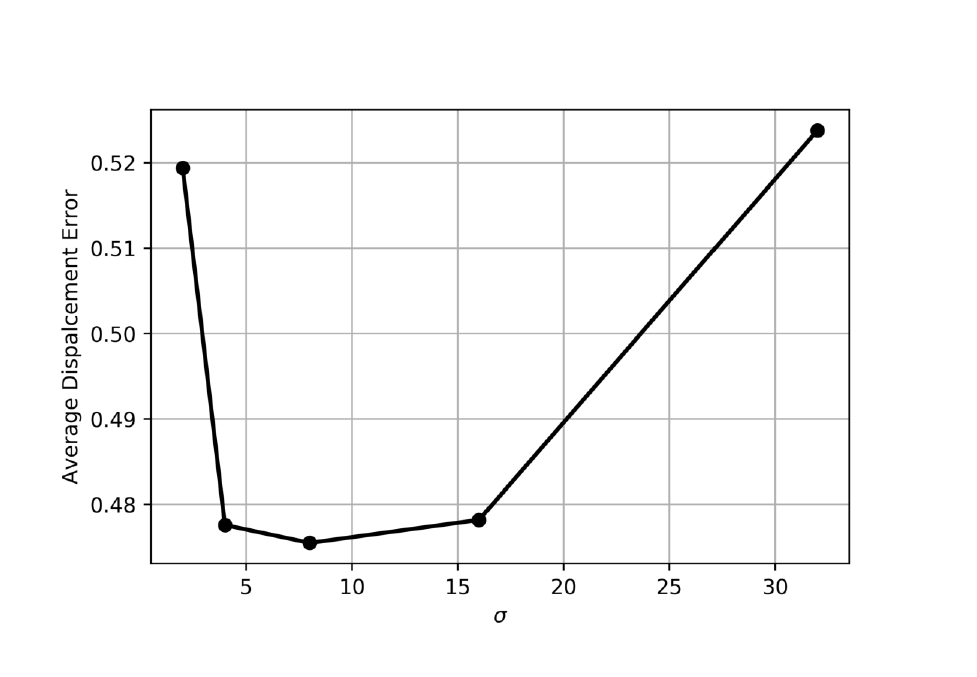}\label{fig8a}}
	\subfloat[FDE]{\includegraphics[width=8cm]{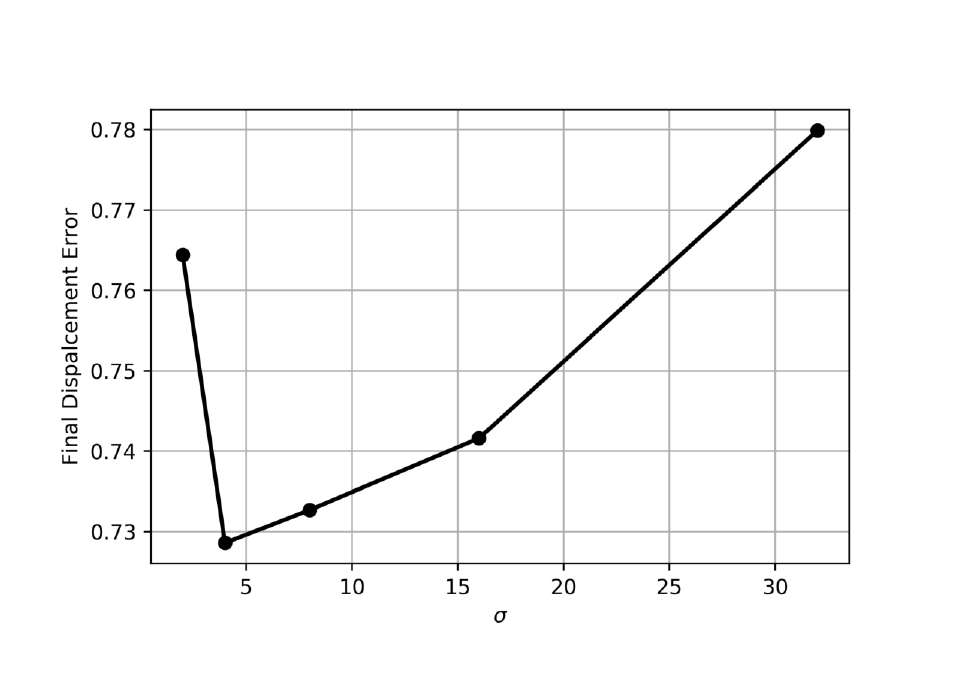}\label{fig8b}}
	\caption{Results of using different $\sigma$.}
	\label{fig8}	
\end{figure*}

\begin{table*}[htbp]
	\caption{Evaluation Results on all the datasets.}
	\begin{center}
		\begin{tabular}{|c|c|c|c|c|c|}
			\hline
			&ETH&Hotel&ZARA01&ZARA02&UCY\\
			\hline
			Number of pedestrians                            &360 &390 &148 &204 &434\\
			
			Number of frames                                 &8603 &11401 &1088 &877 &1352\\
			
			Number of frames within over one pedestrian      &7737 &10516 &1075 &862 &1352\\
			\hline
			
		\end{tabular}
		\label{tab2}
	\end{center}
\end{table*}

\begin{table*}[htbp]
	\caption{Experimental Results on all datasets.}
	\begin{center}
		\resizebox{\textwidth}{!}{
			\begin{tabular}{|c|c|c|c|c|c|c|c|c|} 
				\hline
				Metric&Dataset&Social LSTM/\textit{Var}&Social Attention/\textit{Var}&Ours ($\sigma=2$)/\textit{Var}&Ours ($\sigma=4$)/\textit{Var}&Ours ($\sigma=8$)/\textit{Var}&Ours ($\sigma=16$)/\textit{Var}&Ours ($\sigma=32$)/\textit{Var}\\
				\hline
				\multirow{6}{*}{Average Displacement Error (ADE)} 
				&  ETH                 &  0.5045          /\textit{0.0001} &  0.4834           /\textit{0.0001} &  0.4926          /\textit{0.0001} &  \textbf{0.4285} /\textit{0.0001}  &  0.4321          /\textit{0.0001} &  0.4970          /\textit{0.0001} &  0.4902          /\textit{0.0001}\\ 
				&     Hotel            &  0.5293          /\textit{0.0002} &  0.5423           /\textit{0.0002} &  0.6389          /\textit{0.0002} &  \textbf{0.5044} /\textit{0.0001}  &  0.5536          /\textit{0.0001} &  0.5136          /\textit{0.0002} &  0.5731          /\textit{0.0001}\\ 
				&     ZARA01           &  0.5404          /\textit{0.0032} &  0.6479           /\textit{0.0041} &  0.4930          /\textit{0.0046} &  0.6217          /\textit{0.0036}  &  0.4565          /\textit{0.0033} &  \textbf{0.4382} /\textit{0.0019} &  0.4474          /\textit{0.0008}\\ 
				&     ZARA02           &  0.4827          /\textit{0.0012} &  \textbf{0.2336}  /\textit{0.0002} &  0.4025          /\textit{0.0010} &  0.3639          /\textit{0.0007}  &  0.4159          /\textit{0.0009} &  0.4002          /\textit{0.0013} &  0.5664          /\textit{0.0016}\\ 
				&     UCY              &  0.5653          /\textit{0.0001} &  0.5048           /\textit{0.0001} &  0.5700          /\textit{0.0016} & \textbf{0.4695}  /\textit{0.0001}  &  0.5196          /\textit{0.0001} &  0.5422          /\textit{0.0001} &  0.5421          /\textit{0.0001}\\ 
				&   \textit{Average}   &  0.5244          /\textit{0.00096}&  0.4824           /\textit{0.00096}&  0.5194          /\textit{0.0015} & 0.4776  /\textit{0.00092} &  \textbf{0.4755} /\textit{0.0009} &  0.4782          /\textit{0.00072}&  0.5238          /\textit{0.00054}\\ 
				\hline
				\multirow{6}{*}{Final Displacement Error (FDE)} 
				&     ETH             &  0.9477  /\textit{0.0003} &  0.9581           /\textit{0.0011} &  0.9213          /\textit{0.0005}&  \textbf{0.7679}  /\textit{0.0004}  &  0.8522          /\textit{0.0003} &  0.9297          /\textit{0.0007} &  0.8608          /\textit{0.0004}\\ 
				&     Hotel           &  0.9362  /\textit{0.0011} &  0.8709           /\textit{0.0009} &  0.9635          /\textit{0.0011}&  \textbf{0.7975}  /\textit{0.0006}  &  0.9377          /\textit{0.0010} &  0.8734          /\textit{0.0013} &  0.9489          /\textit{0.0007}\\ 
				&     ZARA01          &  0.9717  /\textit{0.0271} &  1.0644           /\textit{0.0298} &  0.5433          /\textit{0.0316}&  0.8369         /\textit{0.0310}  &  0.5489          /\textit{0.0414} &  0.4604          /\textit{0.0115} & \textbf{0.4213}  /\textit{0.0091}\\ 
				&     ZARA02          &  0.7099  /\textit{0.0050} &  \textbf{0.2486}  /\textit{0.0005} &  0.5334          /\textit{0.0051}&  0.5477           /\textit{0.0031}  &  0.5574          /\textit{0.0028} &  0.5685          /\textit{0.0047} &  0.8710          /\textit{0.0052}\\ 
				&     UCY             &  0.8546  /\textit{0.0004} &  0.7289           /\textit{0.0004} &  0.8606          /\textit{0.0006}&  \textbf{0.6929}  /\textit{0.0004}  &  0.7674          /\textit{0.0008} &  0.8761          /\textit{0.0007} &  0.7978          /\textit{0.0006}\\
				&   \textit{Average}  &  0.8840  /\textit{0.00678}&  0.7742           /\textit{0.00654}&  0.7644          /\textit{0.00778}&  \textbf{0.7286}  /\textit{0.00710} &  0.7327          /\textit{0.00926}&  0.7416          /\textit{0.00378}&  0.7799          /\textit{0.0032}\\ 
				\hline
			\end{tabular}
		}
		\label{tab1}
	\end{center}
\end{table*}

In Equation \eqref{eq9}, $\sigma$ denotes the Gaussian kernel size, namely, the bandwidth related to personal space. According to \cite{ kivinen2004online, Williams2003Learning}, Gaussian kernel size has great anti jamming ability for noise in data, and data are almost useless beyond the $3\times\sigma$ range (99.73\%). In this case, the value of $\sigma$ can represent the radius of one domain where human-human interactions are effective to a certain extent. The value of $\sigma$ determines the amount of domain information the target pedestrian can receive in the scene. If the interaction occurs beyond the effective domain, it would be considered an outlier and excluded from our proposed mechanism. The Gaussian kernel is also able to represent personal space in which the generated interaction is large. The Gaussian kernel has strong nonlinearity and a large value in a certain range, which is similar to the personal space of a pedestrian.
Experiments are conducted to show how different values of $\sigma$ affect the performance of our model. The values of $\sigma$ are selected from 2, 4, 8, 16, and 32, which cover a reasonable range of integer values. 

Experiments are conducted on the aforementioned datasets, and the results are averaged to visualize the performance variations in Figure \ref{fig8}. Table \ref{tab1} provides a comprehensive overview of the outcomes for different values of $\sigma$. Both Figures \ref{fig8a} and \ref{fig8b} exhibit similar trends. Notably, when $\sigma=2$, our model's performance suffers due to limited information shared around the target pedestrian, resulting in minimal human-human interactions considered. Conversely, an excessively large $\sigma$ leads to a performance decrease. This may be attributed to including irrelevant information, such as individuals far from the target pedestrian, in the predictions. As evident from Figures \ref{fig8a} and \ref{fig8b}, our model achieves the minimum Average Displacement Error (ADE) when $\sigma=8$ and the minimum Final Displacement Error (FDE) when $\sigma=4$. Within the appropriate range of $\sigma$ (between 4 and 16), our model consistently achieves competitive results.

\subsection{Quantitative Analysis}

\begin{figure*}[htbp]
	\centering	
	\subfloat[ETH]{\includegraphics[width=6cm]{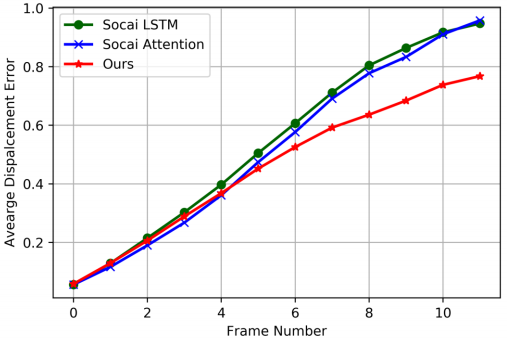}\label{fig6a}}\hfil
	\subfloat[Hotel]{\includegraphics[width=6cm]{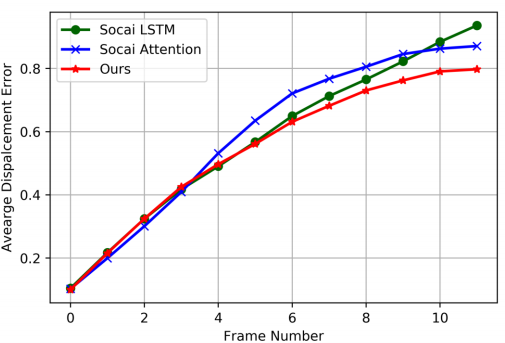}\label{fig6b}}\hfil
	\subfloat[ZARA01]{\includegraphics[width=6cm]{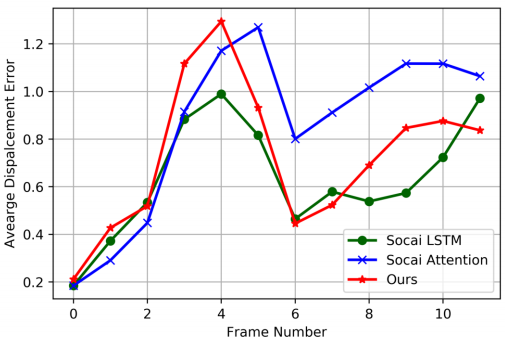}\label{fig6c}}\\ \hfil
	\subfloat[ZARA02]{\includegraphics[width=6cm]{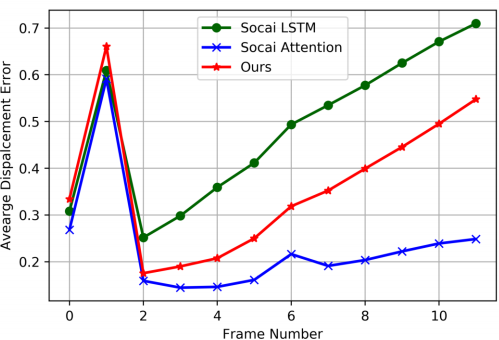}\label{fig6d}}\hfil
	\subfloat[UCY]{\includegraphics[width=6cm]{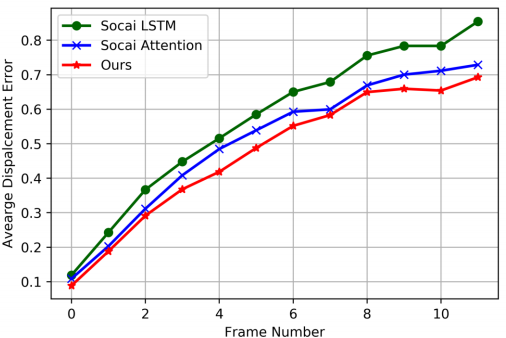}\label{fig6e}}\hfil
	\caption{Variations of the error of each frame. The x-axis represents the frame number to be predicted, ranking from 1 to 12, and the y-axis represents the average displacement error.}
	\label{fig6}	
\end{figure*}

\begin{figure*}[htbp]
	\centering	
	\subfloat[]{\includegraphics[width=8cm, height = 5.3cm]{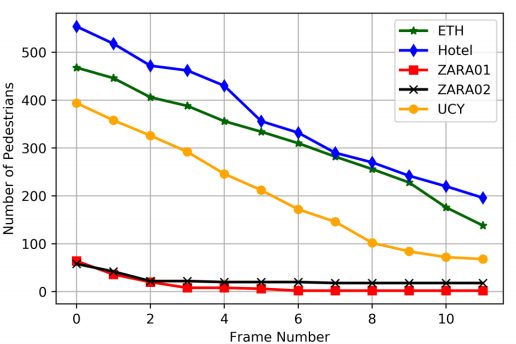}\label{fig7a}}\hfil
	\subfloat[]{\includegraphics[width=8cm, height = 5.3cm]{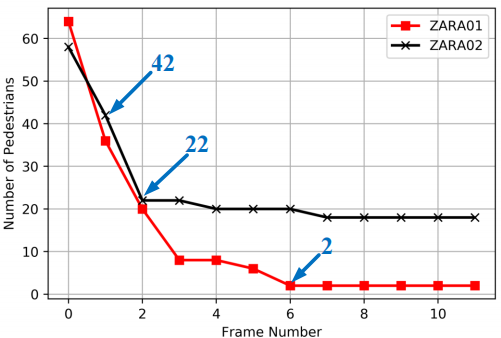}\label{fig7b}}\hfil
	\caption{Variations of the number of pedestrians whose trajectories are predicted.}
	\label{fig7}	
\end{figure*}

First, the complexity of the five scenes is analyzed, and the results are shown in Table \ref{tab2}, where the numbers of pedestrians in each case have large differences. Specifically, less than 150 pedestrians are in case ZARA01, and only 204 pedestrians are in case ZARA02, which means these two cases are relatively sparse. By contrast, cases ETH, Hotel, and UCY consist of more pedestrians. In such crowded cases, human-human interactions would be much more complex and difficult to measure. To present the differences between dissimilar pedestrian cases further, the number of frames within more than one pedestrian where human-human interactions occur is also calculated. The ETH and Hotel datasets contain more crowded frames than the other cases, in which implicit human-human interactions are too complex and difficult to measure.

Table \ref{tab1} shows that in general, our model outperforms the ``Social LSTM" model and the ``Social Attention" model based on two metrics (the results with $\sigma=4$ are taken as examples for analysis). First, in the average ADE and FDE metrics, our model achieves better performance with 8.92\% ADE and 17.56\% FDE improvement compared with the ``Social LSTM" model. The proposed method can provide applicable prediction results to actual pedestrian scenes because these datasets are all collected from the real world. In particular, our model remarkably outperforms two other models on relatively crowded datasets, ETH, Hotel, and UCY. Compared with the “Social LSTM” model in the most crowded UCY case, our model reduces ADE and FDE by 16.95\% and 18.92\%, respectively. Thus, using correntropy to measure human-human interactions is useful and effective. 
While for the ``Social LSTM" model, it only performs better on case ZARA01 than our model with $\sigma=4$ and worse than our model with other values of $\sigma$ according to the ADE metric. As the ``Social LSTM" model assumes all human-human interactions are the same, it cannot balance the different interactions occurring to the same pedestrian and make corresponding predictions, which leads to the highest average prediction ADE and FDE. The ``Social Attention" model achieves the best performance only on case ZARA02, which means the attention mechanism is effective in sparser scenes. When facing complex crowded scenes (ETH, Hotel, and UCY), our model surpasses the ``Social Attention" model. The results prove that our proposed mechanism based on correntropy is more robust for measuring different complex human-human interactions in crowded scenes.

To demonstrate the effectiveness of our model further, the change of average error is plotted along with the frame number in Figure \ref{fig6}. Figures \ref{fig6a}, \ref{fig6b}, and \ref{fig6e} plot the ADE in crowded datasets ETH, Hotel, and UCY. Our model achieves a lower average error than other models, especially in the latter frame numbers (frame number $\geq$ 6). Accuracy in long-term prediction is important in real applications (e.g., autonomous driving and robot navigation). However, the superposition of prediction errors would make achieving accurate long-term prediction difficult. The lower ADE of our method in the long-term prediction confirms that our model could provide applicable prediction results in these crowded datasets.

\begin{figure*}[htbp]	
	\flushleft	
	\subfloat[]{\includegraphics[width=8cm]{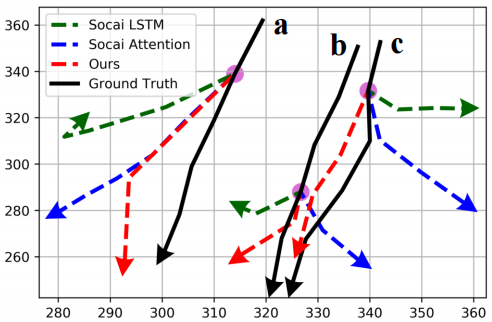}\label{fig5a}}\hfil
	\subfloat[]{\includegraphics[width=8cm]{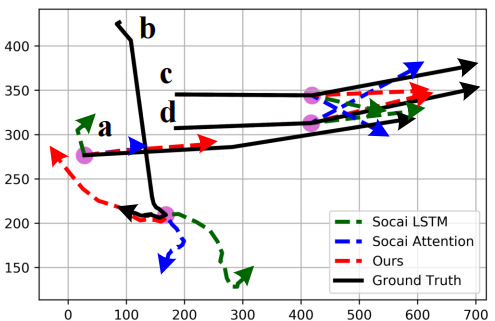}\label{fig5b}}\hfil\\
	\subfloat[]{\includegraphics[width=8cm]{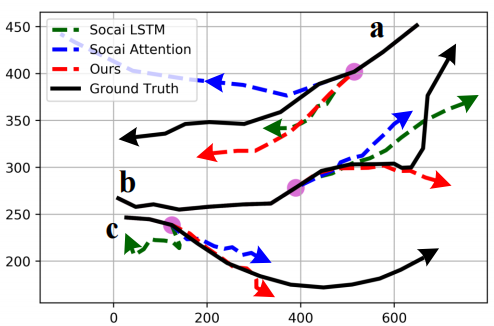}\label{fig5c}}\hfil
	\subfloat[]{\includegraphics[width=8cm]{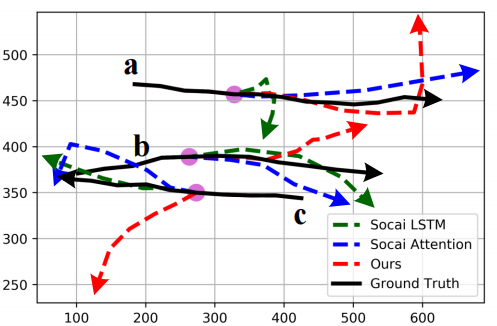}\label{fig5d}}\hfil
	\caption{Illustration of predicted trajectories. These examples contains different complex but common situations in real world.}
	\label{fig5}	
\end{figure*}

In Figures \ref{fig6a}, \ref{fig6b}, and \ref{fig6e}, average error increases when frame number increases, and it is congenial with reason and common sense. However, in Figures \ref{fig6c} and \ref{fig6d}, the change is dramatic, that is, average error increases at the beginning and then plunges. To determine the reason, the number of pedestrians who participate in trajectory prediction in each frame is plotted in Figure \ref{fig7}. Figure \ref{fig7a} shows the number of pedestrians decreases along with the frame in all datasets because the number of pedestrians whose trajectories are longer than 20 frames (8 frames for observation and 12 frames for predictions) is small, especially in relatively sparser cases ZARA01 and ZARA02. Figure \ref{fig7b} is an enlarged illustration of the number change in scenes ZARA01 and ZARA02. In case ZARA01, only two pedestrians participate in prediction because frame 6 and this specific person's trajectory decides the whole error. Thus, the accumulative error before frame 6 disappears, and this situation corresponds to the sudden drop in Figure \ref{fig6c}. In case ZARA02, from frame 1 to frame 2, the number of pedestrians is reduced from 42 to 22. The sudden change of the number greatly decreases the accumulative error, and it is a possible reason for the sudden drop in \ref{fig6d}.

Above all, the whole experimental results prove that the proposed mechanism based on correntropy is able to measure the relative importance of interactions effectively, and our model has the ability to understand different complex human-human interactions in the scene to make applicable prediction results for actual pedestrian cases.

\subsection{Qualitative Analysis}

In this section, we present visualizations of the predicted trajectories and conduct a social analysis of various interactions. Examples of predicted trajectories are illustrated in Figure \ref{fig5}, where pedestrians are labeled with letters (a, b, c, ...) at the initiation of three trajectories, and the purple circle denotes the prediction starting point. It's important to note that the original pedestrian data was normalized in the conducted experiments and subsequently remapped to the real-world scene. The coordinate axis reflects the resolution size of the actual scene. For a more detailed quantitative assessment, we utilize the normalized original data to calculate the Average Displacement Error (ADE) for each pedestrian compared to the ground truth, employing Equation \eqref{eq_ADE}. A lower ADE signifies better predictive accuracy.

In Figure \ref{fig5a}, the predicted trajectories using our model (dashed red line) are closer to the ground truth (solid black line) than other trajectories generated by the ``Social LSTM" model and the ``Social Attention" model because our model could better measure human-human interactions and make accurate predictions. Thus, the ADE of all pedestrians of our model is 0.0149, which is much lower than that of the ``Social LSTM" model (0.0674) and the ``Social Attention" model (0.0505). Specifically, on the right side, two pedestrians (\textbf{b} and \textbf{c}) walk in a group, and one person (pedestrian \textbf{a}) walks alone on the left side. In a social way, as they walk in parallel, they almost have no influence on each other. As correntropy has the ability to measure the feature similarity, our proposed mechanism attaches a relatively small value to human-human interactions and makes accurate predictions. As a result, our model could achieve low ADEs in pedestrian \textbf{b} (0.0139) and \textbf{c} (0.0186). This finding shows that our model can provide applicable prediction results to actual pedestrian trajectories. However, the ``Social LSTM" model (dashed green line) and the ``Social Attention" model (dashed blue line) directly calculate human-human interactions and make predictions, which makes pedestrians want to enlarge the distance among them and leave each other, namely the deviation phenomena. Thus, the ``Social LSTM" model and the ``Social Attention" model achieve larger ADEs in pedestrians \textbf{b} (0.0559/0.0428) and \textbf{c} (0.0817/0.0623).

In the vertical trajectory (pedestrian \textbf{b}) in Figure \ref{fig5b}, the direction of the trajectory predicted by our model is totally opposite to the trajectory generated by the two other models because the ``Social LSTM" model and the ``Social Attention" model incorrectly measure the interaction caused by pedestrian \textbf{a}. Therefore, our model achieves a lower ADE (0.1081) for pedestrian \textbf{b} than the ``Social LSTM" model (0.20123) and the ``Social Attention" model (0.1302). Considering the pedestrians (\textbf{c} and \textbf{d}) who are walking in parallel, the phenomenon is similar to that in Figure \ref{fig5a}, where the ``Social LSTM" model and the ``Social Attention" model make an incorrect prediction of the future positions based on their assumptions. The ADEs of the ``Social LSTM" model and the ``Social Attention" model are 0.1318, 0.1392 for pedestrian \textbf{c} and 0.0779, 0.0764 for pedestrian \textbf{d}. With our proposed mechanism, the relative importance of human-human interactions is correctly measured, and the predicted results are more accurate. The lower ADEs of our model in pedestrians \textbf{c} (0.0699) and \textbf{d} (0.0643) present the effectiveness of the proposed mechanism.

In Figure \ref{fig5c}, three single persons walk in different directions with diverse velocities, and the interactions between them are dissimilar. The ADE of all pedestrians of our model in Figure \ref{fig5c} is 0.1495, and the ADEs of all pedestrians of the ``Social LSTM" model and the ``Social Attention" model are 0.2669 and 0.1722, respectively. The results illustrate that our model can measure the interactions more correctly and make better predictions. According to the middle trajectory (pedestrian \textbf{b}), our model fails to make predictions of the sudden turn and obtains a relatively high ADE metric, 0.1719. Focus in our future work will be on how to make better predictions of ``emergencies" such as sudden turn or sudden stop. 

Figure \ref{fig5d} presents the scene where pedestrians (\textbf{b} and \textbf{c}) walk toward each other with a close distance. Interactions between close pedestrians are complex. In general, when two pedestrians are too close, human-human interactions would be large, and pedestrians would yield right-of-way to avoid collision. However, according to the ground truth, pedestrian \textbf{b} and \textbf{c} do not affect each other possibly because they contain less environmental attention and bring less interactions, which is a special case. Our proposed method learns most of the general situations, so it assigns a high importance to the interaction between pedestrians \textbf{b} and \textbf{c} leading to prediction deviation. That reason is why our model achieves relatively a high ADE metric in pedestrian \textbf{c} (0.1690). Then, the predicted trajectory of pedestrian \textbf{b} leads to a sharp turn of pedestrian \textbf{a} to avoid collision. As a result, the ADE of our model for pedestrian \textbf{a} is 0.0893. The situation of Figure \ref{fig5d} is special, which would not affect the overall performance of the proposed method, and special situations will be further studied in our future work.

\subsection{Comparison with Existing Works}

Our model is compared with several recent existing works: (1) Social LSTM \cite{Alahi2016Social}: This method combines information from all neighboring states by introducing “social” pooling layers. (2) MX-LSTM \cite{hasan2018mx}: This method captures the interplay between tracklets and vislets to forecast the positions and head orientations of an individual. (3) SR-LSTM \cite{zhang2019sr}: This method activates the utilization of the current intention of neighbors and passes the message in the crowd. 

Table \ref{comparison} shows our model outperforms other methods in relatively crowded ETH and UCY datasets, in which human-human interactions are much more complex and difficult to measure. This finding indicates that our proposed mechanism based on correntropy is more robust for measuring different complex human-human interactions in crowded scenes.
In relatively sparse ZARA01 and ZARA02 datasets, our model performs better than the other models on the FDE metrics and achieves competitive results with other models on the ADE metrics. The results prove that our model also has an effect on sparse scenes.

\begin{table}[htbp]
\caption{Comparison with several recent models.}
\label{comparison}
\begin{tabular}{|c|c|c|c|c|}
\hline \multirow{2}{*}{ Method } & \multicolumn{4}{|c|}{ Performance (ADE/FDE) } \\
\cline { 2 - 5 } & ETH & UCY & ZARA01 & ZARA02 \\
\hline Social LSTM & $0.50/0.94$ & $0.57/0.85$ & $0.54/0.97$ & $0.48/0.71$ \\
MX-LSTM & $-/-$ & $0.49/1.12$ & $0.59/1.31$ & $0.35/0.79$ \\
SR-LSTM & $0.63/1.25$ & $0.51/1.10$ & $\textbf{0.41}/0.90$ & $\textbf{0.32}/0.70$ \\
\hline OURs & $\textbf{0.43}/\textbf{0.77}$ & $\textbf{0.48}/\textbf{0.73}$ & $0.44\textbf{/0.46}$ & $0.36/\textbf{0.55}$  \\
\hline
\end{tabular}
\end{table}

\subsection{Model Speed Analysis}

To elaborate the model speed, NVIDIA 3090 GPU is used to train our model and calculate the inference time by averaging 100 times of single inference steps. Specifically, our model achieves a quick pedestrian trajectory prediction at a speed of 0.049 s per inference step. This result means that our model is computationally efficient and meets the online real-time computing requirements of autonomous driving and mobile robot. Our model achieves such quick inference speed because the designed ``Interaction Module” can autonomously capture different complex human-human interactions among crowds directly from the original data. It does not introduce additional notations or social rules.
    
\section{Conclusion}
In this study, we introduce the Interaction-Aware LSTM (IA-LSTM) model designed for predicting pedestrians' future trajectories.

Our model excels in gauging the relative importance of diverse human–human interactions within a scene and subsequently shares weighted feature representations through the "Interaction Module" to enhance trajectory predictions. Experimental results showcase that our proposed model outperforms several state-of-the-art methods across two public datasets featuring five scenes, particularly excelling in crowded scenarios. Moreover, our qualitative analysis highlights the model's success in predicting various behaviors stemming from human–human interactions in a socially plausible manner, including group walking and parallel movement.

Future endeavors will involve extending our approach to diverse scenes and enhancing the "Interaction Module" to predict emergencies or sudden turns effectively.



\section*{Acknowledgments}
This work was supported by the National Key Research and Development Program of China under Grant No. 2020AAA0108100, the National Natural Science Foundation of China under Grant No. 62073257, 62141223, and the Key Research and Development Program of Shaanxi Province of China under Grant No. 2022GY-076.

%





\ifCLASSOPTIONcaptionsoff
  \newpage
\fi

\bibliographystyle{IEEEtran}
\bibliography{IEEEtranrefs}

\begin{thebibliography}{100}
\providecommand{\url}[1]{#1}
\csname url@samestyle\endcsname
\providecommand{\newblock}{\relax}
\providecommand{\bibinfo}[2]{#2}
\providecommand{\BIBentrySTDinterwordspacing}{\spaceskip=0pt\relax}
\providecommand{\BIBentryALTinterwordstretchfactor}{4}
\providecommand{\BIBentryALTinterwordspacing}{\spaceskip=\fontdimen2\font plus
\BIBentryALTinterwordstretchfactor\fontdimen3\font minus \fontdimen4\font\relax}
\providecommand{\BIBforeignlanguage}[2]{{%
\expandafter\ifx\csname l@#1\endcsname\relax
\typeout{** WARNING: IEEEtran.bst: No hyphenation pattern has been}%
\typeout{** loaded for the language `#1'. Using the pattern for}%
\typeout{** the default language instead.}%
\else
\language=\csname l@#1\endcsname
\fi
#2}}
\providecommand{\BIBdecl}{\relax}
\BIBdecl

\bibitem{Lerner2010Crowds}
A.~Lerner, Y.~Chrysanthou, and D.~Lischinski, ``Crowds by example,'' \emph{Computer Graphics Forum}, vol.~26, no.~3, pp. 655--664, 2010.

\bibitem{Luo2018PORCA}
Y.~Luo, P.~Cai, A.~Bera, D.~Hsu, W.~S. Lee, and D.~Manocha, ``Porca: Modeling and planning for autonomous driving among many pedestrians,'' \emph{IEEE Robotics \& Automation Letters}, vol.~PP, no.~99, pp. 1--1, 2018.

\bibitem{Pellegrini2009You}
S.~Pellegrini, A.~Ess, K.~Schindler, and L.~J.~V. Gool, ``You'll never walk alone: Modeling social behavior for multi-target tracking,'' in \emph{IEEE International Conference on Computer Vision}, 2009, pp. 261--268.

\bibitem{Vivacqua2017Self}
R.~P.~D. Vivacqua, M.~Bertozzi, P.~Cerri, F.~N. Martins, and R.~F. Vassallo, ``Self-localization based on visual lane marking maps: An accurate low-cost approach for autonomous driving,'' \emph{IEEE Transactions on Intelligent Transportation Systems}, vol.~PP, no.~99, pp. 1--16, 2017.

\bibitem{Thrun2002MINERVA}
S.~Thrun, M.~Bennewitz, W.~Burgard, A.~B. Cremers, F.~Dellaert, D.~Fox, D.~Hahnel, C.~Rosenberg, N.~Roy, and J.~Schulte, ``Minerva: A second-generation museum tour-guide robot,'' in \emph{IEEE International Conference on Robotics \& Automation}, vol.~3, 2002, pp. 1999--2005.

\bibitem{Leonard1990Application}
J.~J. Leonard and H.~F. Durrant-Whyte, ``Application of multi-target tracking to sonar-based mobile robot navigation,'' in \emph{IEEE Conference on Decision \& Control}, vol.~29, 1990, pp. 3118--3123.

\bibitem{Trautman2010Unfreezing}
P.~Trautman and A.~Krause, ``Unfreezing the robot: Navigation in dense, interacting crowds,'' in \emph{IEEE/RSJ International Conference on Intelligent Robots \& Systems}, 2010, pp. 797--803.

\bibitem{8540942}
Z.~Wan, C.~Jiang, M.~Fahad, Z.~Ni, Y.~Guo, and H.~He, ``Robot-assisted pedestrian regulation based on deep reinforcement learning,'' \emph{IEEE Transactions on Cybernetics}, vol.~50, no.~4, pp. 1669--1682, 2020.

\bibitem{8694838}
Y.~Feng, Y.~Yuan, and X.~Lu, ``Person reidentification via unsupervised cross-view metric learning,'' \emph{IEEE Transactions on Cybernetics}, vol.~51, no.~4, pp. 1849--1859, 2021.

\bibitem{target_tracking_CYB1}
X.~Cao, L.~Ren, and C.~Sun, ``Dynamic target tracking control of autonomous underwater vehicle based on trajectory prediction,'' \emph{IEEE Transactions on Cybernetics}, pp. 1--14, 2022.

\bibitem{target_tracking_CYB2}
P.~Li, S.~Wang, H.~Yang, and H.~Zhao, ``Trajectory tracking and obstacle avoidance for wheeled mobile robots based on empc with an adaptive prediction horizon,'' \emph{IEEE Transactions on Cybernetics}, pp. 1--10, 2021.

\bibitem{shafiee2021introvert}
N.~Shafiee, T.~Padir, and E.~Elhamifar, ``Introvert: Human trajectory prediction via conditional 3d attention,'' in \emph{Proceedings of the IEEE/CVF Conference on Computer Vision and Pattern Recognition}, 2021, pp. 16\,815--16\,825.

\bibitem{moussaid2010walking}
M.~Moussa{\"\i}d, N.~Perozo, S.~Garnier, D.~Helbing, and G.~Theraulaz, ``The walking behaviour of pedestrian social groups and its impact on crowd dynamics,'' \emph{PloS one}, vol.~5, no.~4, p. e10047, 2010.

\bibitem{gupta2018social}
A.~Gupta, J.~Johnson, L.~Feifei, S.~Savarese, and A.~Alahi, ``Social gan: Socially acceptable trajectories with generative adversarial networks,'' in \emph{Computer Vision \& Pattern Recognition}, 2018, pp. 2255--2264.

\bibitem{helbing1995social}
D.~Helbing and P.~Molnar, ``Social force model for pedestrian dynamics,'' \emph{Physical Review E}, vol.~51, no.~5, pp. 4282--4286, 1995.

\bibitem{yang2019top}
D.~Yang, L.~Li, K.~Redmill, and {\"U}.~{\"O}zg{\"u}ner, ``Top-view trajectories: A pedestrian dataset of vehicle-crowd interaction from controlled experiments and crowded campus,'' in \emph{2019 IEEE Intelligent Vehicles Symposium (IV)}.\hskip 1em plus 0.5em minus 0.4em\relax IEEE, 2019, pp. 899--904.

\bibitem{yamaguchi2011you}
K.~Yamaguchi, A.~C. Berg, L.~E. Ortiz, and T.~L. Berg, ``Who are you with and where are you going?'' in \emph{CVPR 2011}.\hskip 1em plus 0.5em minus 0.4em\relax IEEE, 2011, pp. 1345--1352.

\bibitem{Alahi2016Social}
A.~Alahi, K.~Goel, V.~Ramanathan, A.~Robicquet, and S.~Savarese, ``Social lstm: Human trajectory prediction in crowded spaces,'' in \emph{Computer Vision \& Pattern Recognition}, 2016, pp. 961--971.

\bibitem{hochreiter1997long}
S.~Hochreiter and J.~Schmidhuber, ``Long short-term memory,'' \emph{Neural Computation}, vol.~9, no.~8, pp. 1735--1780, 1997.

\bibitem{chung2014empirical}
J.~Chung, C.~Gulcehre, K.~Cho, and Y.~Bengio, ``Empirical evaluation of gated recurrent neural networks on sequence modeling,'' \emph{arXiv: Neural and Evolutionary Computing}, vol. 1412.3555, 2014.

\bibitem{Fernando2018Tree}
T.~Fernando, S.~Denman, A.~Mcfadyen, S.~Sridharan, and C.~Fookes, ``Tree memory networks for modelling long-term temporal dependencies,'' \emph{Neurocomputing}, vol. 304, pp. 64--81, 2018.

\bibitem{Kitani2011Fast}
K.~M. Kitani, T.~Okabe, Y.~Sato, and A.~Sugimoto, ``Fast unsupervised ego-action learning for first-person sports videos,'' in \emph{Computer Vision \& Pattern Recognition}, 2011, pp. 3241--3248.

\bibitem{Ryoo2015Early}
M.~S. Ryoo, T.~J. Fuchs, L.~Xia, J.~K. Aggarwal, and L.~Matthies, ``Early recognition of human activities from first-person videos using onset representations,'' in \emph{Computer Vision \& Pattern Recognition}, 2015.

\bibitem{Srivastava2015Unsupervised}
N.~Srivastava, E.~Mansimov, and R.~Salakhutdinov, ``Unsupervised learning of video representations using lstms,'' in \emph{International Conference on Machine Learning}, 2015, pp. 843--852.

\bibitem{vondrick2015anticipating}
C.~Vondrick, H.~Pirsiavash, and A.~Torralba, ``Anticipating the future by watching unlabeled video.'' \emph{Computer Vision \& Pattern Recognition}, 2015.

\bibitem{vemula2018social}
A.~Vemula, K.~Muelling, and J.~Oh, ``Social attention: Modeling attention in human crowds,'' in \emph{International Conference on Robotics and Automation}, 2018, pp. 1--7.

\bibitem{vaswani2017attention}
A.~Vaswani, N.~Shazeer, N.~Parmar, J.~Uszkoreit, L.~Jones, A.~N. Gomez, L.~Kaiser, and I.~Polosukhin, ``Attention is all you need,'' \emph{Neural Information Processing Systems}, pp. 5998--6008, 2017.

\bibitem{graves2014towards}
A.~Graves and N.~Jaitly, ``Towards end-to-end speech recognition with recurrent neural networks,'' in \emph{International conference on machine learning}.\hskip 1em plus 0.5em minus 0.4em\relax PMLR, 2014, pp. 1764--1772.

\bibitem{graves2013generating}
A.~Graves, ``Generating sequences with recurrent neural networks,'' \emph{arXiv preprint arXiv:1308.0850}, 2013.

\bibitem{wu2023multi}
Y.~Wu, L.~Wang, S.~Zhou, J.~Duan, G.~Hua, and W.~Tang, ``Multi-stream representation learning for pedestrian trajectory prediction,'' in \emph{Proceedings of the AAAI Conference on Artificial Intelligence}, vol.~37, no.~3, 2023, pp. 2875--2882.

\bibitem{shi2023trajectory}
L.~Shi, L.~Wang, S.~Zhou, and G.~Hua, ``Trajectory unified transformer for pedestrian trajectory prediction,'' in \emph{Proceedings of the IEEE/CVF International Conference on Computer Vision}, 2023, pp. 9675--9684.

\bibitem{zhou2023static}
H.~Zhou, X.~Yang, M.~Fan, H.~Huang, D.~Ren, and H.~Xia, ``Static-dynamic global graph representation for pedestrian trajectory prediction,'' \emph{Knowledge-Based Systems}, vol. 277, p. 110775, 2023.

\bibitem{personal_space1}
N.~Bhargava, S.~Chaudhuri, and G.~Seetharaman, ``Linear cyclic pursuit based prediction of personal space violation in surveillance video,'' in \emph{2013 IEEE Applied Imagery Pattern Recognition Workshop (AIPR)}, 2013, pp. 1--5.

\bibitem{personal_space2}
T.~Li, H.~Chang, M.~Wang, B.~Ni, R.~Hong, and S.~Yan, ``Crowded scene analysis: A survey,'' \emph{IEEE Transactions on Circuits and Systems for Video Technology}, vol.~25, no.~3, pp. 367--386, 2015.

\bibitem{personal_space3}
A.~Sardar, M.~Joosse, A.~Weiss, and V.~Evers, ``Don't stand so close to me: Users' attitudinal and behavioral responses to personal space invasion by robots,'' in \emph{2012 7th ACM/IEEE International Conference on Human-Robot Interaction (HRI)}, 2012, pp. 229--230.

\bibitem{personal_space4}
Z.~Bingchen, G.~Weimin, Z.~Hengyu, C.~Huachun, and W.~Yanqun, ``Research on relationship of passenger's behavior and space,'' in \emph{2011 IEEE 2nd International Conference on Computing, Control and Industrial Engineering}, vol.~1, 2011, pp. 82--85.

\bibitem{golchoubian2023pedestrian}
M.~Golchoubian, M.~Ghafurian, K.~Dautenhahn, and N.~L. Azad, ``Pedestrian trajectory prediction in pedestrian-vehicle mixed environments: A systematic review,'' \emph{IEEE Transactions on Intelligent Transportation Systems}, 2023.

\bibitem{zhang2023dual}
X.~Zhang, P.~Angeloudis, and Y.~Demiris, ``Dual-branch spatio-temporal graph neural networks for pedestrian trajectory prediction,'' \emph{Pattern Recognition}, vol. 142, p. 109633, 2023.

\bibitem{10103218}
X.~Zhong, X.~Yan, Z.~Yang, W.~Huang, K.~Jiang, R.~W. Liu, and Z.~Wang, ``Visual exposes you: Pedestrian trajectory prediction meets visual intention,'' \emph{IEEE Transactions on Intelligent Transportation Systems}, vol.~24, no.~9, pp. 9390--9400, 2023.

\bibitem{pellegrini2010improving}
S.~Pellegrini, A.~Ess, and L.~Van~Gool, ``Improving data association by joint modeling of pedestrian trajectories and groupings,'' in \emph{European Conference on Computer Vision}, vol. 6311, 2010, pp. 452--465.

\bibitem{Schneider2013Pedestrian}
N.~Schneider and D.~M. Gavrila, ``Pedestrian path prediction with recursive bayesian filters: A comparative study,'' in \emph{German Conference on Pattern Recognition}, vol. 8142, 2013, pp. 174--183.

\bibitem{Ballan2016Knowledge}
L.~Ballan, F.~Castaldo, A.~Alahi, F.~Palmieri, and S.~Savarese, ``Knowledge transfer for scene-specific motion prediction,'' in \emph{European Conference on Computer Vision}, 2016, pp. 697--713.

\bibitem{kooij2014context-based}
J.~F.~P. Kooij, N.~Schneider, F.~Flohr, and D.~M. Gavrila, ``Context-based pedestrian path prediction,'' \emph{European Conference on Computer Vision}, vol. 8694, pp. 618--633, 2014.

\bibitem{huang2016deep}
S.~Huang, X.~Li, Z.~Zhang, Z.~He, F.~Wu, W.~Liu, J.~Tang, and Y.~Zhuang, ``Deep learning driven visual path prediction from a single image,'' \emph{IEEE Transactions on Image Processing}, vol.~25, no.~12, pp. 5892--5904, 2016.

\bibitem{walker2014patch}
J.~Walker, A.~Gupta, and M.~Hebert, ``Patch to the future: Unsupervised visual prediction,'' in \emph{Computer Vision \& Pattern Recognition}, 2014, pp. 2224--2231.

\bibitem{Dan2013Inferring}
X.~Dan, S.~Todorovic, and S.~C. Zhu, ``Inferring "dark matter" and "dark energy" from videos,'' in \emph{IEEE International Conference on Computer Vision}, 2013, pp. 2224--2231.

\bibitem{Shuai2016Pedestrian}
Y.~Shuai, H.~Li, and X.~Wang, ``Pedestrian behavior understanding and prediction with deep neural networks,'' in \emph{European Conference on Computer Vision}, 2016, pp. 263--279.

\bibitem{Abbeel2016Inverse}
W.~Xue, B.~Lian, J.~Fan, P.~Kolaric, T.~Chai, and F.~L. Lewis, ``Inverse reinforcement q-learning through expert imitation for discrete-time systems,'' \emph{IEEE Transactions on Neural Networks and Learning Systems}, pp. 1--14, 2021.

\bibitem{Ng2000Algorithms}
A.~Y. Ng and S.~Russell, ``Algorithms for inverse reinforcement learning,'' in \emph{International Conference on Machine Learning}, vol.~67, no.~2, 2000, pp. 663--670.

\bibitem{Ziebart2009Planning}
B.~D. Ziebart, N.~D. Ratliff, G.~Gallagher, C.~Mertz, K.~M. Peterson, J.~A. Bagnell, M.~Hebert, A.~K. Dey, and S.~S. Srinivasa, ``Planning-based prediction for pedestrians,'' in \emph{IEEE/RSJ International Conference on Intelligent Robots \& Systems}, 2009, pp. 3931--3936.

\bibitem{Kitani2012Activity}
K.~M. Kitani, B.~D. Ziebart, J.~A. Bagnell, and M.~Hebert, ``Activity forecasting,'' in \emph{European Conference on Computer Vision}, 2012, pp. 201--214.

\bibitem{Lee2016Predicting}
N.~Lee and K.~M. Kitani, ``Predicting wide receiver trajectories in american football,'' in \emph{Applications of Computer Vision}, 2016, pp. 1--9.

\bibitem{ma2017forecasting}
W.~Ma, D.~Huang, N.~Lee, and K.~M. Kitani, ``Forecasting interactive dynamics of pedestrians with fictitious play,'' in \emph{Computer Vision \& Pattern Recognition}, 2017, pp. 4636--4644.

\bibitem{bae2023set}
I.~Bae and H.-G. Jeon, ``A set of control points conditioned pedestrian trajectory prediction,'' in \emph{Proceedings of the AAAI Conference on Artificial Intelligence}, vol.~37, no.~5, 2023, pp. 6155--6165.

\bibitem{liang2023stglow}
R.~Liang, Y.~Li, J.~Zhou, and X.~Li, ``Stglow: a flow-based generative framework with dual-graphormer for pedestrian trajectory prediction,'' \emph{IEEE transactions on neural networks and learning systems}, 2023.

\bibitem{CYB_2023_1}
X.~Cao, L.~Ren, and C.~Sun, ``Dynamic target tracking control of autonomous underwater vehicle based on trajectory prediction,'' \emph{IEEE Transactions on Cybernetics}, vol.~53, no.~3, pp. 1968--1981, 2023.

\bibitem{CYB_2023_2}
X.~Na and D.~J. Cole, ``Experimental evaluation of a game-theoretic human driver steering control model,'' \emph{IEEE Transactions on Cybernetics}, vol.~53, no.~8, pp. 4791--4804, 2023.

\bibitem{CYB_2023_3}
P.~Li, S.~Wang, H.~Yang, and H.~Zhao, ``Trajectory tracking and obstacle avoidance for wheeled mobile robots based on empc with an adaptive prediction horizon,'' \emph{IEEE Transactions on Cybernetics}, vol.~52, no.~12, pp. 13\,536--13\,545, 2022.

\bibitem{CYB_2023_4}
H.~Gao, H.~An, W.~Lin, X.~Yu, and J.~Qiu, ``Trajectory tracking of variable centroid objects based on fusion of vision and force perception,'' \emph{IEEE Transactions on Cybernetics}, vol.~53, no.~12, pp. 7957--7965, 2023.

\bibitem{CYB_2023_5}
Z.~Xiao, H.~Fang, H.~Jiang, J.~Bai, V.~Havyarimana, H.~Chen, and L.~Jiao, ``Understanding private car aggregation effect via spatio-temporal analysis of trajectory data,'' \emph{IEEE Transactions on Cybernetics}, vol.~53, no.~4, pp. 2346--2357, 2023.

\bibitem{Liu2007Correntropy}
W.~Liu, P.~P. Pokharel, and J.~C. Principe, ``Correntropy: Properties and applications in non-gaussian signal processing,'' \emph{IEEE Transactions on Signal Processing}, vol.~55, no.~11, pp. 5286--5298, 2007.

\bibitem{Principe:2010:ITL:1855180}
J.~C. Principe, \emph{Information Theoretic Learning: Renyi's Entropy and Kernel Perspectives}, 1st~ed.\hskip 1em plus 0.5em minus 0.4em\relax Springer Publishing Company, Incorporated, 2010.

\bibitem{Santamar2006Generalized}
I.~Santamaría, P.~P. Pokharel, and J.~C. Principe, ``Generalized correlation function: Definition, properties, and application to blind equalization,'' \emph{IEEE Transactions on Signal Processing}, vol.~54, no.~6, pp. 2187--2197, 2006.

\bibitem{Principe2010Information}
J.~Príncipe, ``Information theoretic learning,'' in \emph{International Workshop on Pattern Recognition in Information Systems}, 2010.

\bibitem{ShiTraining}
W.~Shi, Y.~Gong, X.~Tao, and N.~Zheng, ``Training dcnn by combining max-margin, max-correlation objectives, and correntropy loss for multilabel image classification,'' \emph{IEEE Transactions on Neural Networks and Learning Systems}, vol.~29, no.~7, pp. 2896--2908, 2018.

\bibitem{Singh2014The}
A.~Singh, R.~Pokharel, and J.~Principe, ``The c-loss function for pattern classification,'' \emph{Pattern Recognition}, vol.~47, no.~1, pp. 441--453, 2014.

\bibitem{Syed2012Correntropy}
N.~M. Syed, J.~Principe, and P.~Pardalos, ``Correntropy in data classification,'' \emph{Springer Proceedings in Mathematics and Statistics}, vol.~20, pp. 81--117, 01 2012.

\bibitem{Chen2012Recursive}
X.~Chen, Y.~Jian, J.~Liang, and Q.~Ye, ``Recursive robust least squares support vector regression based on maximum correntropy criterion,'' \emph{Neurocomputing}, vol.~97, no. Complete, pp. 63--73, 2012.

\bibitem{Feng2015Learning}
Y.~Feng, X.~Huang, S.~Lei, Y.~Yang, and J.~A.~K. Suykens, ``Learning with the maximum correntropy criterion induced losses for regression,'' \emph{Journal of Machine Learning Research}, vol.~16, no.~1, pp. 993--1034, 2015.

\bibitem{Chen2016Efficient}
L.~Chen, Q.~Hua, J.~Zhao, B.~Chen, and J.~C. Principe, ``Efficient and robust deep learning with correntropy-induced loss function,'' \emph{Neural Computing \& Applications}, vol.~27, no.~4, pp. 1019--1031, 2016.

\bibitem{Xu2008A}
J.~Xu and J.~C. Principe, ``A pitch detector based on a generalized correlation function,'' \emph{IEEE Transactions on Audio, Speech, and Language Processing}, vol.~16, no.~8, pp. 1420--1432, 2008.

\bibitem{Chen2014Steady}
B.~Chen, X.~Lei, J.~Liang, N.~Zheng, and J.~C. Principe, ``Steady-state mean-square error analysis for adaptive filtering under the maximum correntropy criterion,'' \emph{IEEE Signal Processing Letters}, vol.~21, no.~7, pp. 880--884, 2014.

\bibitem{Zhao2011Kernel}
S.~Zhao, B.~Chen, and J.~C. Principe, ``Kernel adaptive filtering with maximum correntropy criterion,'' in \emph{International Joint Conference on Neural Networks}, 2011, pp. 2012--2017.

\bibitem{Chen2018Robust}
B.~Chen, L.~Xing, X.~Wang, J.~Qin, and N.~Zheng, ``Robust learning with kernel mean $p$-power error loss,'' \emph{IEEE Transactions on Cybernetics}, vol.~PP, no.~99, pp. 1--13, 2018.

\bibitem{Ran2011Robust}
H.~Ran, H.~Bao-Gang, Z.~Wei-Shi, and K.~Xiang-Wei, ``Robust principal component analysis based on maximum correntropy criterion,'' \emph{IEEE Transactions on Image Processing}, vol.~20, no.~6, pp. 1485--1494, 2011.

\bibitem{chen2021multikernel}
B.~Chen, Y.~Xie, X.~Wang, Z.~Yuan, P.~Ren, and J.~Qin, ``Multikernel correntropy for robust learning,'' \emph{IEEE Transactions on Cybernetics}, 2021.

\bibitem{8303753}
B.~Du, T.~Xinyao, Z.~Wang, L.~Zhang, and D.~Tao, ``Robust graph-based semisupervised learning for noisy labeled data via maximum correntropy criterion,'' \emph{IEEE Transactions on Cybernetics}, vol.~49, no.~4, pp. 1440--1453, 2019.

\bibitem{7447745}
Y.~Wang, Y.~Y. Tang, and L.~Li, ``Correntropy matching pursuit with application to robust digit and face recognition,'' \emph{IEEE Transactions on Cybernetics}, vol.~47, no.~6, pp. 1354--1366, 2017.

\bibitem{8959408}
K.~Xiong, H.~H.~C. Iu, and S.~Wang, ``Kernel correntropy conjugate gradient algorithms based on half-quadratic optimization,'' \emph{IEEE Transactions on Cybernetics}, vol.~51, no.~11, pp. 5497--5510, 2021.

\bibitem{M1995BTT}
T.~Yang and L.~O. Chua, ``Implementing back-propagation- through-time learning algorithm using cellular neural networks,'' \emph{International Journal of Bifurcation \& Chaos}, 1999.

\bibitem{Williams1989Experimental}
R.~J. Williams and D.~Zipser, ``Experimental analysis of the real-time recurrent learning algorithm,'' \emph{Connection Science}, vol.~1, no.~1, pp. 87--111, 1989.

\bibitem{mikolov2010recurrent}
T.~Mikolov, M.~Karafiat, L.~Burget, J.~Cernocký, and S.~Khudanpur, ``Recurrent neural network based language model,'' pp. 1045--1048, 2010.

\bibitem{Chorowski2014End}
J.~Chorowski, D.~Bahdanau, K.~Cho, and Y.~Bengio, ``End-to-end continuous speech recognition using attention-based recurrent nn: First results,'' \emph{ArXiv: Neural and Evolutionary Computing}, 2014.

\bibitem{chung2015a}
J.~Chung, K.~Kastner, L.~Dinh, K.~Goel, A.~C. Courville, and Y.~Bengio, ``A recurrent latent variable model for sequential data,'' \emph{Neural Information Processing Systems}, pp. 2980--2988, 2015.

\bibitem{Graves2013Speech}
A.~Graves, A.~Mohamed, and G.~Hinton, ``Speech recognition with deep recurrent neural networks,'' in \emph{IEEE International Conference on Acoustics, Speech, and Signal Processing}, 2013, pp. 6645--6649.

\bibitem{young2017recent}
T.~Young, D.~Hazarika, S.~Poria, and E.~Cambria, ``Recent trends in deep learning based natural language processing.'' \emph{ArXiv: Computation and Language}, 2017.

\bibitem{LSTM1}
S.~Z. Tajalli, A.~Kavousi-Fard, M.~Mardaneh, A.~Khosravi, and R.~Razavi-Far, ``Uncertainty-aware management of smart grids using cloud-based lstm-prediction interval,'' \emph{IEEE Transactions on Cybernetics}, pp. 1--14, 2021.

\bibitem{LSTM2}
X.~Xu and M.~Yoneda, ``Multitask air-quality prediction based on lstm-autoencoder model,'' \emph{IEEE Transactions on Cybernetics}, vol.~51, no.~5, pp. 2577--2586, 2021.

\bibitem{LSTM3}
Y.~Han, W.~Qi, N.~Ding, and Z.~Geng, ``Short-time wavelet entropy integrating improved lstm for fault diagnosis of modular multilevel converter,'' \emph{IEEE Transactions on Cybernetics}, pp. 1--9, 2021.

\bibitem{denoord2016pixel}
A.~V. Den~Oord, N.~Kalchbrenner, and K.~Kavukcuoglu, ``Pixel recurrent neural networks,'' in \emph{International Conference on Machine Learning}, 2016, pp. 1747--1756.

\bibitem{Karpathy2015Deep}
A.~Karpathy and F.~F. Li, ``Deep visual-semantic alignments for generating image descriptions,'' in \emph{Computer Vision \& Pattern Recognition}, 2015, pp. 3128--3137.

\bibitem{LSTM_CV}
M.~Liu, H.~Hu, L.~Li, Y.~Yu, and W.~Guan, ``Chinese image caption generation via visual attention and topic modeling,'' \emph{IEEE Transactions on Cybernetics}, pp. 1--11, 2020.

\bibitem{LSTM4}
O.~Wu, T.~Yang, M.~Li, and M.~Li, ``Two-level lstm for sentiment analysis with lexicon embedding and polar flipping,'' \emph{IEEE Transactions on Cybernetics}, pp. 1--13, 2020.

\bibitem{sutskever2014sequence}
I.~Sutskever, O.~Vinyals, and Q.~V. Le, ``Sequence to sequence learning with neural networks,'' \emph{neural information processing systems}, pp. 3104--3112, 2014.

\bibitem{Niccol2018Group}
N.~Bisagno, B.~Zhang, and N.~Conci, ``Group lstm: Group trajectory prediction in crowded scenarios,'' \emph{Springer, Cham}, 2018.

\bibitem{zhang2019sr}
P.~Zhang, W.~Ouyang, P.~Zhang, J.~Xue, and N.~Zheng, ``Sr-lstm: State refinement for lstm towards pedestrian trajectory prediction,'' in \emph{Proceedings of the IEEE/CVF Conference on Computer Vision and Pattern Recognition}, 2019, pp. 12\,085--12\,094.

\bibitem{2020CF}
Y.~Xu, J.~Yang, and S.~Du, ``Cf-lstm: Cascaded feature-based long short-term networks for predicting pedestrian trajectory.'' in \emph{National Conference on Artificial Intelligence}, 2020.

\bibitem{Luber2010People}
M.~Luber, J.~A. Stork, G.~D. Tipaldi, and O.~A. Kai, ``People tracking with human motion predictions from social forces,'' in \emph{IEEE International Conference on Robotics \& Automation}, 2010, pp. 464--469.

\bibitem{Mehran2009Abnormal}
R.~Mehran, A.~Oyama, and M.~Shah, ``Abnormal crowd behavior detection using social force model,'' in \emph{IEEE Conference on Computer Vision \& Pattern Recognition}, 2009, pp. 935--942.

\bibitem{9044739}
Y.~Zhu, H.~Zhao, X.~Zeng, and B.~Chen, ``Robust generalized maximum correntropy criterion algorithms for active noise control,'' \emph{IEEE/ACM Transactions on Audio, Speech, and Language Processing}, vol.~28, pp. 1282--1292, 2020.

\bibitem{9511644}
L.~Gao, X.~Li, D.~Bi, L.~Peng, X.~Xie, and Y.~Xie, ``Robust tensor recovery in impulsive noise based on correntropy and hybrid tensor sparsity,'' \emph{IEEE Transactions on Circuits and Systems II: Express Briefs}, pp. 1--1, 2021.

\bibitem{9475523}
H.~Zhao, D.~Liu, and S.~Lv, ``Robust maximum correntropy criterion subband adaptive filter algorithm for impulsive noise and noisy input,'' \emph{IEEE Transactions on Circuits and Systems II: Express Briefs}, pp. 1--1, 2021.

\bibitem{9115431}
W.~Shi and Y.~Li, ``A shrinkage correntropy based algorithm under impulsive noise environment,'' in \emph{2019 6th International Conference on Information, Cybernetics, and Computational Social Systems (ICCSS)}, 2019, pp. 244--247.

\bibitem{lealtaixe2014learning}
L.~Lealtaixe, M.~Fenzi, A.~Kuznetsova, B.~Rosenhahn, and S.~Savarese, ``Learning an image-based motion context for multiple people tracking,'' in \emph{Computer Vision \& Pattern Recognition}, 2014, pp. 3542--3549.

\bibitem{kivinen2004online}
J.~Kivinen, A.~J. Smola, and R.~C. Williamson, ``Online learning with kernels,'' \emph{IEEE Transactions on Signal Processing}, vol.~52, no.~8, pp. 2165--2176, 2004.

\bibitem{Williams2003Learning}
C.~K.~I. Williams, ``Learning with kernels: Support vector machines, regularization, optimization, and beyond,'' \emph{Publications of the American Statistical Association}, vol.~98, no. 462, pp. 489--489, 2003.

\bibitem{hasan2018mx}
I.~Hasan, F.~Setti, T.~Tsesmelis, A.~Del~Bue, F.~Galasso, and M.~Cristani, ``Mx-lstm: mixing tracklets and vislets to jointly forecast trajectories and head poses,'' in \emph{Proceedings of the IEEE Conference on Computer Vision and Pattern Recognition}, 2018, pp. 6067--6076.

\end{thebibliography}

\begin{IEEEbiography}[{\includegraphics[width=1in,height=1.25in,clip,keepaspectratio]{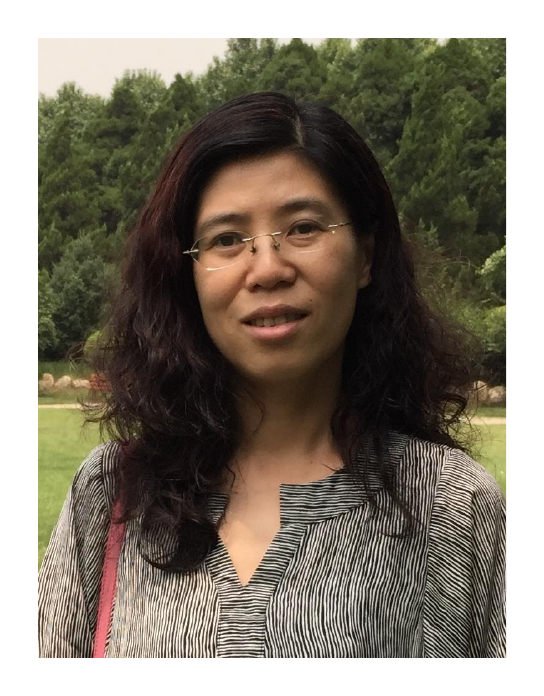}}]{Jing Yang}
received the B.S. and M.S. degrees in control science and engineering and the Ph.D. degree in pattern recognition and intelligent	systems from Xi’an Jiaotong University, China, in 1999, 2002, and 2010, respectively.

From 1999 to 2003, she was a Research Assistant with the Institute of Automation, Xi’an Jiaotong University. Since 2019, she has been an Associate Professor with the Department of Automation Science and Technology, Xi'an Jiaotong University. Her research interests include machine learning, reinforcement learning, and information theory and their applications to intelligent systems such as autonomous vehicles. Since 2004, she has been a member of Intelligent Vehicles Team, Institute of Artificial Intelligence and Robotics, Xi’an Jiaotong University.
\end{IEEEbiography}

\begin{IEEEbiography}[{\includegraphics[width=1in,height=1.25in,clip,keepaspectratio]{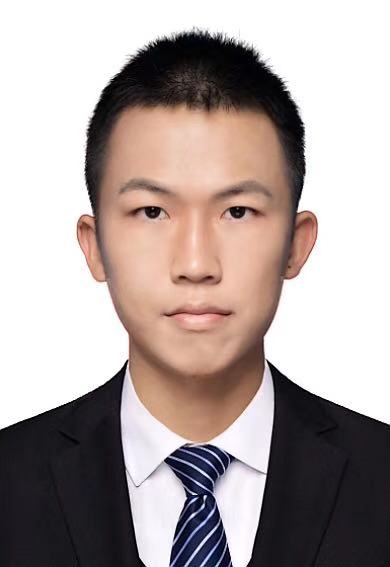}}]{Yuehai Chen}
received the B.S. degree in automation from Xi’an Jiaotong University, Xi’an, Shaanxi, China, in 2020, where he is currently pursuing the doctor’s degree in control science and engineering. His research interests include machine learning, artificial intelligence, computer vision, and their applications to intelligent systems. He is a student member of the IEEE.
\end{IEEEbiography}

\begin{IEEEbiography}[{\includegraphics[width=1in,height=1.25in,clip,keepaspectratio]{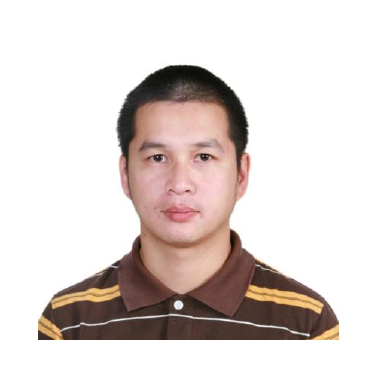}}]{Shaoyi Du}
received the B.S. degrees in computational mathematics and in computer science, the M.S. degree in applied mathematics, and the Ph.D. degree in pattern recognition and intelligence system from Xi’an Jiaotong University, China, in 2002, 2005, and 2009 respectively. 

He was a Post-Doctoral Fellow with Xi’an Jiaotong University from 2009 to 2011 and was with The University of North Carolina at Chapel Hill from 2013 to 2014. He is currently a Professor with the Institute of Artificial Intelligence and Robotics, Xi’an Jiaotong University. His research interests include computer vision, machine learning, and pattern recognition
\end{IEEEbiography}

\begin{IEEEbiography}[{\includegraphics[width=1in,height=1.25in,clip,keepaspectratio]{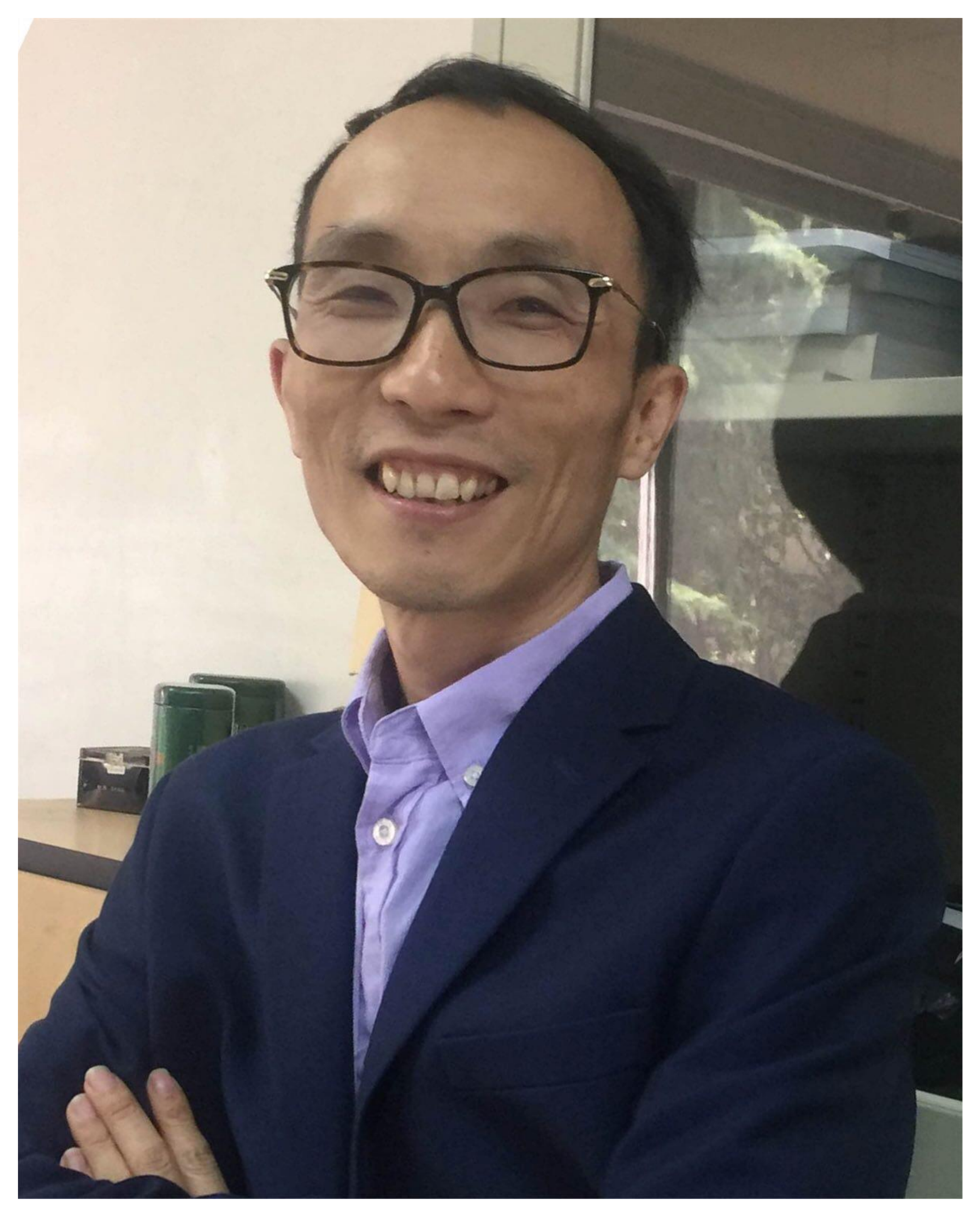}}]{Badong Chen}
received the B.S. and M.S. degrees in control theory and engineering from Chongqing University, in 1997 and 2003, respectively, and the Ph.D. degree in computer science and technology from Tsinghua University in 2008. He was a Postdoctoral Researcher with Tsinghua University from 2008 to 2010, and a Postdoctoral Associate at the University of Florida Computational NeuroEngineering Laboratory (CNEL) during the period October, 2010 to September, 2012. During July to August 2015, he visited the Nanyang Technological University (NTU) as a visiting research scientist. He also served as a senior research fellow with The Hong Kong Polytechnic University from August to November in 2017. Currently he is a professor at the Institute of Artificial Intelligence and Robotics (IAIR), Xi’an Jiaotong University. His research interests are in signal processing, information theory, machine learning, and their applications to cognitive science and neural engineering. He has published 2 books, 4 chapters, and over 200 papers in various journals and conference proceedings. Dr. Chen is an IEEE Senior Member, a Technical Committee Member of IEEE SPS Machine Learning for Signal Processing (MLSP) and IEEE CIS Cognitive and Developmental Systems (CDS), and an associate editor of IEEE Transactions on Cognitive and Developmental Systems, IEEE Transactions on Neural Networks and Learning Systems and Journal of The Franklin Institute, and has been on the editorial board of Entropy.
\end{IEEEbiography}

\begin{IEEEbiography}[{\includegraphics[width=1in,height=1.25in,clip,keepaspectratio]{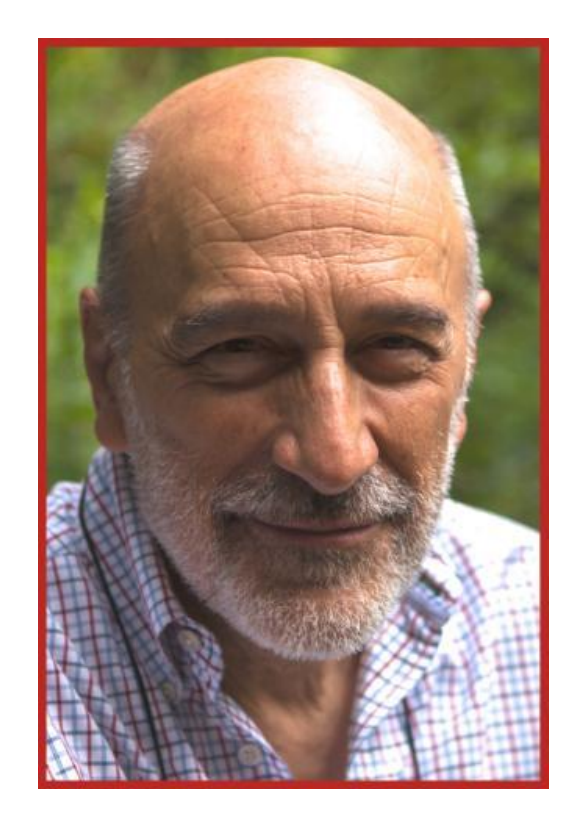}}]{Jose C. Principe}
received the B.S. degree from the University of Porto, Portugal, in 1972 and the M.Sc. and Ph.D. degrees from the University of Florida in 1974 and 1979, respectively. He is a Distinguished Professor of Electrical and Biomedical Engineering with the University of Florida, Gainesville, where he teaches advanced signal processing and artificial neural networks (ANNs) modeling. He is a BellSouth Professor and
Founder and Director of the University of Florida Computational NeuroEngineering Laboratory (CNEL). He is involved in biomedical signal processing, in particular, the electroencephalogram (EEG) and the modeling and applications of adaptive systems. He has more than 129 publications in refereed journals, 15 book chapters, and over 300 conference papers. He has directed more than 50 Ph.D.dissertations and 61 master’s degree theses.

Dr. Principe is Editor-in-Chief of the IEEE TRANSACTIONS ON BIOMEDICAL ENGINEERING, President of the International Neural Network Society, and formal Secretary of the Technical Committee on Neural Networks of the IEEE Signal Processing Society. He is an AIMBE Fellow and a recipient of the IEEE Engineering in Medicine and Biology Society Career Service Award. He is also a member of the Scientific Board of the Food and Drug Administration, and a member of the Advisory Board of the McKnight Brain Institute at the University of Florida.
\end{IEEEbiography}





\end{document}